\documentclass[11pt, a4paper, logo, copyright]{arxiv/googledeepmind}
\makeatletter
\let\oldnewcounter\newcounter
\let\oldsetcounter\setcounter
\renewcommand{\newcounter}[1]{}  
\renewcommand{\setcounter}[2]{}
\usepackage{tocloft}
\let\newcounter\oldnewcounter    
\let\setcounter\oldsetcounter
\makeatother

\usepackage{amsmath}   
\usepackage{amssymb}   
\usepackage{mathtools} 
\usepackage[utf8]{inputenc} 
\usepackage[T1]{fontenc}    
\usepackage{url}            
\usepackage{booktabs}       
\usepackage{amsfonts}       
\usepackage{nicefrac}       
\usepackage{microtype}      
\usepackage{xcolor}         
\usepackage{graphicx}
\usepackage{xspace} 
\usepackage[super]{nth}
\usepackage{subfigure}
\usepackage{colortbl}
\usepackage{wrapfig,booktabs}
\usepackage{multirow}  
\usepackage{caption}  
\usepackage{floatflt}

\usepackage{algorithm}
\usepackage{algpseudocode}
\usepackage{amsmath}
\usepackage{amssymb}
\usepackage{enumitem}
\usepackage{bm} 
\usepackage{lipsum}
\usepackage{standalone}
\usepackage{array} 

\usepackage{gensymb}

\newcommand{\grayrow}{\rowcolor[gray]{.9}}

\newcommand{\projectname}{\normalsize \texttt{AIM}\xspace}

\newcommand{\Largeprojectname}{\Large \texttt{AIM}\xspace}

\definecolor{textgreen}{RGB}{57, 172, 57}
\definecolor{textred}{RGB}{200, 10, 10}
\usepackage{pifont} 
\newcommand{\cmark}{\textcolor{textgreen}{\Large \ding{51}}}%
\newcommand{\xmark}{\textcolor{textred}{\Large  \ding{55}}}%
\usepackage[outline]{contour}
\definecolor{statistical}{HTML}{8c564b}
\definecolor{structural}{HTML}{0070C0}
\definecolor{semantic}{HTML}{008080}
\definecolor{yellow}{HTML}{f7c600}
\definecolor{lightblue}{HTML}{0071bc}
\definecolor{lightgreen}{HTML}{39b54a}
\definecolor{deemph}{gray}{0.55}

\usepackage{xcolor,colortbl}

\definecolor{Gray}{gray}{0.95}
\definecolor{LightCyan}{rgb}{0.88,1,1}
\colorlet{darkgreen}{green!65!black}
\colorlet{darkblue}{blue!75!black}
\colorlet{darkred}{red!80!black}
\definecolor{lightblue}{HTML}{0071bc}
\definecolor{lightgreen}{HTML}{39b54a}
\definecolor{manyshot}{HTML}{6969ff}
\definecolor{medshot}{HTML}{f7c600}
\definecolor{fewshot}{HTML}{ff6969}
\definecolor{mypurple}{HTML}{412F8A}
\definecolor{myorange}{HTML}{fc8e62}

\definecolor{deemph}{gray}{0.55}

\newcolumntype{a}{>{\columncolor{Gray}}r}
\newcolumntype{b}{>{\columncolor{white}}r}

\usepackage{adjustbox,float}
\usepackage{array,multirow,graphicx}

\newcolumntype{R}[2]{%
    >{\adjustbox{angle=#1}\bgroup}%
    l%
    <{\egroup}%
}

\newcommand{\walkIcon}[1][0.92]{
    \hspace*{-0.3cm}
    \includegraphics[width=0.3cm]{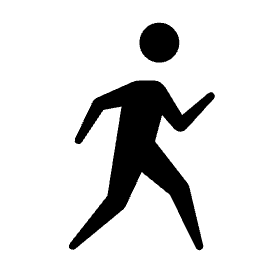}
    \hspace*{-0.2cm}
}

\newcommand{\bikeIcon}[1][0.92]{
    \hspace*{-0.2cm}
    \includegraphics[width=0.3cm]{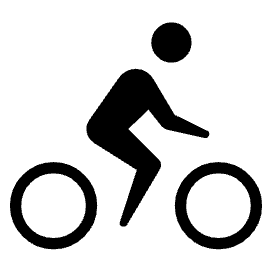}
    \hspace*{-0.2cm}
}

\newcommand{\spinningIcon}[1][0.92]{
    \hspace*{-0.2cm}
    \includegraphics[width=0.3cm]{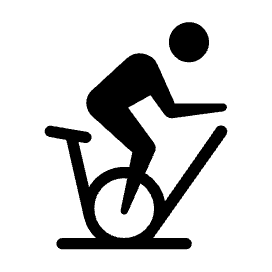}
    \hspace*{-0.2cm}
}

\newcommand{\golfIcon}[1][0.92]{
    \hspace*{-0.3cm}
    \includegraphics[width=0.3cm]{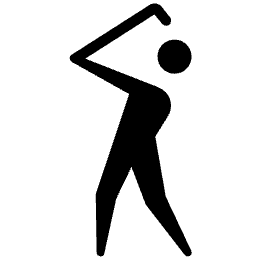}
    \hspace*{-0.2cm}
}

\newcommand{\tennisIcon}[1][0.92]{
    \hspace*{-0.3cm}
    \includegraphics[width=0.3cm]{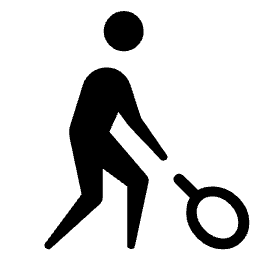}
    \hspace*{-0.2cm}
}

\newcommand{\hikingIcon}[1][0.92]{
    \hspace*{-0.3cm}
    \includegraphics[width=0.3cm]{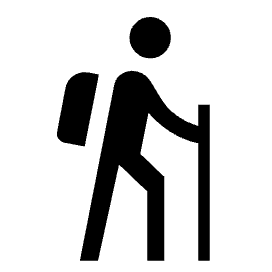}
    \hspace*{-0.2cm}
}

\newcommand{\swimIcon}[1][0.92]{
    \hspace*{-0.2cm}
    \includegraphics[width=0.3cm]{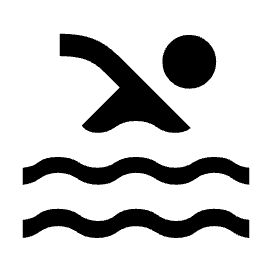}
    \hspace*{-0.2cm}
}

\newcommand{\crossfitIcon}[1][0.92]{
    \hspace*{-0.2cm}
    \includegraphics[width=0.3cm]{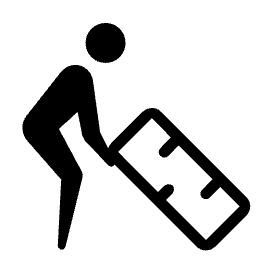}
    \hspace*{-0.2cm}
}

\newcommand{\rollerbladdingIcon}[1][0.92]{
    \hspace*{-0.3cm}
    \includegraphics[width=0.3cm]{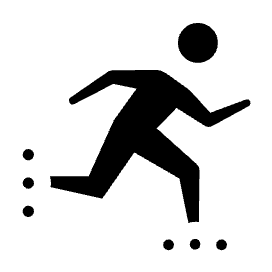}
    \hspace*{-0.2cm}
}

\newcommand{\climbingIcon}[1][0.92]{
    \hspace*{-0.3cm}
    \includegraphics[width=0.3cm]{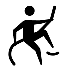}
    \hspace*{-0.2cm}
}

\newcommand{\dancingIcon}[1][0.92]{
    \hspace*{-0.3cm}
    \includegraphics[width=0.3cm]{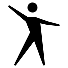}
    \hspace*{-0.2cm}
}

\newcommand{\ellipticalIcon}[1][0.92]{
    \hspace*{-0.2cm}
    \includegraphics[width=0.3cm]{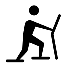}
    \hspace*{-0.2cm}
}

\newcommand{\aerobicsIcon}[1][0.92]{
    \hspace*{-0.3cm}
    \includegraphics[width=0.3cm]{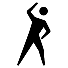}
    \hspace*{-0.2cm}
}

\newcommand{\runningIcon}[1][0.92]{
    \hspace*{-0.3cm}
    \includegraphics[width=0.3cm]{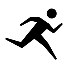}
    \hspace*{-0.2cm}
}

\newcommand{\hiitIcon}[1][0.92]{
    \hspace*{-0.3cm}
    \includegraphics[width=0.3cm]{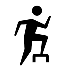}
    \hspace*{-0.2cm}
}

\newcommand{\yogaIcon}[1][0.92]{
    \hspace*{-0.3cm}
    \includegraphics[width=0.3cm]{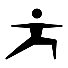}
    \hspace*{-0.2cm}
}

\newcommand{\weightliftingIcon}[1][0.92]{
    \hspace*{-0.3cm}
    \includegraphics[width=0.3cm]{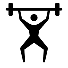}
    \hspace*{-0.2cm}
}

\newcommand{\treadmillIcon}[1][0.92]{
    \hspace*{-0.3cm}
    \includegraphics[width=0.3cm]{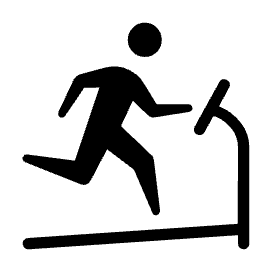}
    \hspace*{-0.2cm}
}

\newcommand{\rollerbladingIcon}[1][0.92]{
    \hspace*{-0.3cm}
    \includegraphics[width=0.3cm]{Icons/fitbit_rollerblading_68dp_000000_FILL0_ROND50_wght400_GRAD0_opsz48.png}
    \hspace*{-0.2cm}
}

\newcommand{\indoorrowingIcon}[1][0.92]{
    \hspace*{-0.3cm}
    \includegraphics[width=0.3cm]{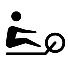}
    \hspace*{-0.2cm}
}

\newcommand{\kickboxingIcon}[1][0.92]{
    \hspace*{-0.3cm}
    \includegraphics[width=0.3cm]{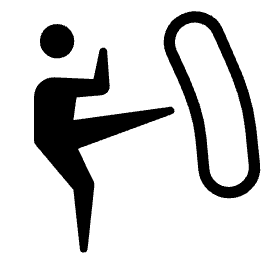}
    \hspace*{-0.2cm}
}

\newcommand{\skiingIcon}[1][0.92]{
    \hspace*{-0.3cm}
    \includegraphics[width=0.3cm]{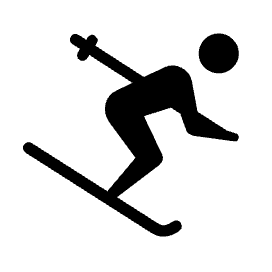}
    \hspace*{-0.2cm}
}

\newcommand{\pilatesIcon}[1][0.92]{
    \hspace*{-0.2cm}
    \includegraphics[width=0.3cm]{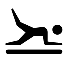}
    \hspace*{-0.2cm}
}

\newcommand{\stairclimberIcon}[1][0.92]{
    \hspace*{-0.2cm}
    \includegraphics[width=0.3cm]{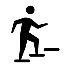}
    \hspace*{-0.2cm}
}

\newcommand{\bootcampIcon}[1][0.92]{
    \hspace*{-0.3cm}
    \includegraphics[width=0.3cm]{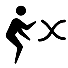}
    \hspace*{-0.2cm}
}

\newcommand{\coretrainingIcon}[1][0.92]{
    \hspace*{-0.3cm}
    \includegraphics[width=0.3cm]{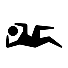}
    \hspace*{-0.2cm}
}

\newcommand{\sportsIcon}[1][0.92]{
    \hspace*{-0.3cm}
    \includegraphics[width=0.3cm]{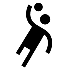}
    \hspace*{-0.2cm}
}

\newcommand{\martialartsIcon}[1][0.92]{
    \hspace*{-0.3cm}
    \includegraphics[width=0.3cm]{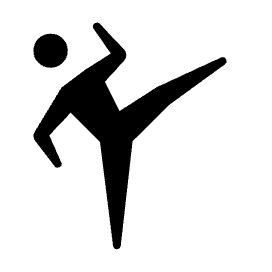}
    \hspace*{-0.2cm}
}

\newcommand{\snowboardingIcon}[1][0.92]{
    \hspace*{-0.3cm}
    \includegraphics[width=0.3cm]{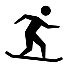}
    \hspace*{-0.2cm}
}

\newcommand{\kayakingIcon}[1][0.92]{
    \hspace*{-0.2cm}
    \includegraphics[width=0.3cm]{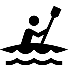}
    \hspace*{-0.2cm}
}

\newcommand{\crosscountryskiingIcon}[1][0.92]{
    \hspace*{-0.3cm}
    \includegraphics[width=0.3cm]{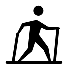}
    \hspace*{-0.2cm}
}

\newcommand{\surfingIcon}[1][0.92]{
    \hspace*{-0.3cm}
    \includegraphics[width=0.3cm]{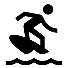}
    \hspace*{-0.2cm}
}

\newcommand{\paddleboardingIcon}[1][0.92]{
    \hspace*{-0.3cm}
    \includegraphics[width=0.3cm]{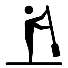}
    \hspace*{-0.2cm}
}

\title{LSM-2: Learning from Incomplete \\Wearable Sensor Data}

\author{
Maxwell A. Xu\textsuperscript{1,3*,$\bm{\dagger}$},
Girish Narayanswamy\textsuperscript{1,4*,$\bm{\dagger}$},
Kumar Ayush\textsuperscript{1},
Dimitris Spathis\textsuperscript{1},
Shun Liao\textsuperscript{1},
Shyam A. Tailor\textsuperscript{1},
Ahmed Metwally\textsuperscript{1},
A. Ali Heydari\textsuperscript{1},
Yuwei Zhang\textsuperscript{1},
Jake Garrison\textsuperscript{1},
Samy Abdel-Ghaffar\textsuperscript{1},
Xuhai Xu\textsuperscript{1},
Ken Gu\textsuperscript{1},
Jacob Sunshine\textsuperscript{1},
Ming-Zher Poh\textsuperscript{1},
Yun Liu\textsuperscript{1},
Tim Althoff\textsuperscript{1},
Shrikanth Narayanan\textsuperscript{2},
Pushmeet Kohli\textsuperscript{2},
Mark Malhotra\textsuperscript{1},
Shwetak Patel\textsuperscript{1},
Yuzhe Yang\textsuperscript{1},
James M. Rehg\textsuperscript{3},
Xin Liu\textsuperscript{1,$\circ$},
Daniel McDuff\textsuperscript{1,$\circ$} \\
\vspace{0.5em}
\normalfont\fontsize{8}{10}\selectfont
\textsuperscript{1}Google Research, 
\textsuperscript{2}Google DeepMind,
\textsuperscript{3}University of Illinois Urbana-Champaign, 
\textsuperscript{4}University of Washington \\
\textsuperscript{$\dagger$}Co-first authors, 
\textsuperscript{$\circ$}Co-last authors, 
\textsuperscript{*}Work done during an internship at Google
}

\correspondingauthor{
  maxu@illinois.edu, 
  girishvn@uw.edu, 
  \{xliucs,dmcduff\}@google.com
}

\begin{abstract}
Foundation models, a cornerstone of recent advancements in machine learning, have predominantly thrived on complete and well-structured data. Wearable sensor data frequently suffers from significant missingness, posing a substantial challenge for self-supervised learning (SSL) models that typically assume complete data inputs. This paper introduces the second generation of Large Sensor Model (LSM-2) with Adaptive and Inherited Masking (\projectname), a novel SSL approach that learns robust representations directly from incomplete data without requiring explicit imputation. \projectname's core novelty lies in its use of learnable mask tokens to model both existing ("inherited") and artificially introduced missingness, enabling it to robustly handle fragmented real-world data during inference. Pre-trained on an extensive dataset of 40M hours of day-long multimodal sensor data, our LSM-2 with  \projectname achieves the best performance across a diverse range of tasks, including classification, regression and generative modeling. Furthermore, LSM-2 with \projectname exhibits superior scaling performance, and critically, maintains high performance even under targeted missingness scenarios, reflecting clinically coherent patterns, such as the diagnostics value of nighttime biosignals for hypertension prediction. This makes \projectname more reliable choice for real-world wearable data applications.
\end{abstract}

\begin{document}

\maketitle
\newcommand{\nocontentsline}[3]{}
\let\origcontentsline\addcontentsline
\newcommand\stoptoc{\let\addcontentsline\nocontentsline}
\newcommand\resumetoc{\let\addcontentsline\origcontentsline}

\stoptoc
\section{Introduction}

In the real world, missing or incomplete data is a pervasive challenge across a variety of domains. In clinical settings for example, electronic health records frequently exhibit missingness due to factors such as loss to follow-up \citep{haneuse2021assessing, zhou2023missing} or condition-specific diagnostic procedures \citep{ford2020can, mcdermott2021comprehensive}. Similarly, sensor systems grapple with incomplete data streams due to strategic intermittent deactivation for energy conservation, environmental noise, sensor obstruction, or hardware malfunctions \citep{du2020missing, bahr2022missing, decorte2024missing}. Missing data for wearable mobile health sensors is especially prevalent and problematic. In addition to the aforementioned causes, user compliance issues (e.g. improper/insecure device attachment) or mobile-specific challenges (e.g. data transmission failures, battery charging periods) further exacerbate the problem \cite{Rahman2017, xu2022pulseimpute}.

Self-Supervised Learning (SSL) has emerged as a powerful paradigm for learning transferable representations by exploiting inherent structures within unlabeled data \citep{ericsson2021well}. When scaled to large pre-training datasets with sufficient compute, these approaches yield foundation models capable of strong generalization across diverse downstream tasks \citep{oquab2023dinov2, team2023gemini}. This is especially promising for wearable sensors, where physiological signals contain rich information predictive of diverse health outcomes, with several recent large-scale data collection initiatives, such as UK Biobank \citep{Katori2022}, All of Us \citep{Jeong2025allofus}, and the Apple Heart and Movement Study \citep{truslow2024understanding}. This has enabled the development of wearable sensor foundation models that generalize across multiple health prediction tasks \citep{narayanswamy2024scaling, xu2024relcon,saha2025pulse, abbaspourazad2023large}.

\begin{wrapfigure}{b!}{0.5\textwidth}
    \vspace{-4mm}
    \centering
    \includegraphics[width=.5\textwidth]{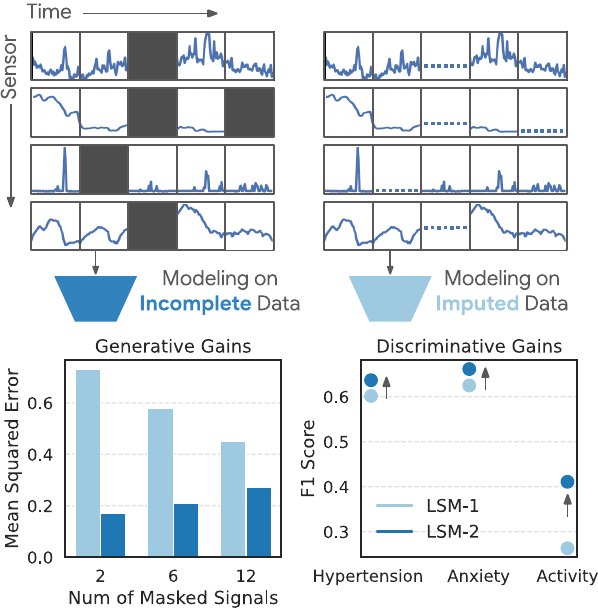}
    \caption{\textbf{LSM-2 Models Incomplete Data.} Our method uses a learned mask token to represent existing missingness during inference. Then, if sensors are missing, it can directly reconstruct them [L] or classify directly on the incomplete data [R].}
    \vspace{-10pt}
    \label{fig:teaser} 
\end{wrapfigure}

Unfortunately, state-of-the-art time-series SSL approaches typically assume fully-observed data inputs. As such, prior wearable sensor foundation models have handled missingness by modeling short context windows (i.e. <60s \citep{abbaspourazad2023large}, 2.56s \citep{xu2024relcon}, 10s \citep{pillai2024papagei}), where incomplete instances can easily be filtered out. However, many clinically relevant physiological patterns (e.g. circadian rhythms \citep{zielinski2014strengths}, heart rate variability \citep{chuduc2013review}, and daily activity profiles \citep{hecht2009methodology}) require analyzing day-long recordings. Unfortunately, day-long data inevitably contains missingness due to wearable sensor limitations (e.g. battery drain necessitating strategic sensor deactivation, motion artifacts corrupting signals). As detailed in Section~\ref{sec:datadescription}, our dataset exhibits pervasive missingness: 0\% of records are complete. While prior work with similar data employed imputation methods in order to train their SSL model \citep{narayanswamy2024scaling}, such approaches risk introducing biases that can propagate to downstream models~\citep{jungo2024representation}.

In this paper, we introduce the second generation of Large Sensor Model (LSM-2) based on \underline{A}daptive and \underline{I}nherited \underline{M}asking, {\large\texttt{AIM}}, a self-supervised learning approach that learns a representation directly from incomplete data with diverse missingness patterns. To the best of our knowledge, this is the first work to address representation learning directly on incomplete wearable sensor data.  Building on masked autoencoder (MAE) pre-training \citep{he2022maskedmae}, \projectname uses a shared learnable mask token to represent both inherited and artificial masks. \emph{Inherited masks} are derived from existing missingness in raw data, thereby masking incomplete data and avoiding the need for imputation. \emph{Artificial masks}, are randomly applied on observed tokens, providing a ground truth for the reconstruction pre-training objective. Via \projectname's introduction of inherited masks, mask tokens are learned to represent real-world missingness. During evaluation, missingness still occurs in the raw data. Here, the inherited mask allows for missingness-aware embeddings. Like real missingness, the number of inherited mask tokens may vary, violating the naive MAE's assumption of a fixed number of masked tokens~\citep{he2022maskedmae}. As such, the \emph{adaptive} component of \projectname is able to suppress any additional missing tokens from contributing to the final encoder output, ensuring that the encoding is a learned representation of the non-missing data solely. This encoding can then be used in conjunction with a linear probe to predict a variety of downstream classification and regression tasks, as well as being fed back into the decoder for downstream generative tasks.

\textbf{The key contributions of our work are:}
\vspace{-3mm}
\begin{enumerate}[leftmargin=*,noitemsep]
    \item  \textit{We introduce LSM-2 and propose a novel training strategy, \underline{A}daptive and \underline{I}nherited \underline{M}asking,} {\large\texttt{AIM}}, that uses adaptive masking to jointly model artificial and inherited missingness and learn a strong, generalizable representation, directly on incomplete data. By incorporating adaptive masking during pre-training and inference, our method enables a single model to robustly support a variety of downstream tasks under real-world missingness conditions without requiring any explicit imputation.
    \item \textit{We demonstrate that our LSM-2 w/} \projectname \textit{pre-trained foundation model achieves state-of-the-art performance} across diverse set of tasks (3x classification, 4x generative, 3x regression) that cover a wide range of semantics (clinical, mental health, wearables, demographics) after large-scale pre-training on 40 million hours of day-long multimodal sensor data. Our model also demonstrates superior scaling performance as compared to our prior LSM-1 model ~\citep{narayanswamy2024scaling}.
    \item \textit{We evaluate the robustness of our LSM-2 across a wide range of targeted missing scenarios}, dropping out specific sensors or time windows, and we demonstrate much less performance degradation compared to the baseline method that is pre-trained with imputed data. The missingness scenarios in which our model does express sensitivity is reflective of physiological domain knowledge, providing interesting insights into the nature of a given prediction target. 
\end{enumerate}

\section{Related Work}
\label{sec:background}

\textbf{Self-Supervised Learning for Time-Series Foundation Models.} Our LSM-2 model utilizes \projectname, an MAE \citep{he2022maskedmae} SSL framework that combines an artificial mask with an inherited mask from real-world sensor data. This differs from LSM-1 \citep{narayanswamy2024scaling}, the most closely-related work, which performs MAE pre-training with just an artificial mask and uses naive imputation to fill in pre-existing missingness, both of which negatively impacts downstream performance (see Section~\ref{sec:resultsanddiscussion}). Other MAE-style methods for time series data are limited in that they either:
(a) focus exclusively on complete univariate signals~\citep{dong2023simmtm, li2023ti, chien2022maeeg}, (b) work with highly correlated channels from a single modality~\citep{na2024guiding}, or (c) focus on task-specific forecasting without learning general-purpose embeddings~\citep{ansari2024chronos, nie2022time, das2024decoder}. Notably, none of these approaches handle the missing data patterns inherent in real-world multivariate sensor data. Alternatively, contrastive SSL methods learn representations by attracting positives and repelling negatives in embedding space. Positives are generated via augmentations \citep{tang2020exploring} or sampling using temporal proximity \citep{tonekaboni2021unsupervised}, subject labels \citep{abbaspourazad2023large}, domain knowledge \citep{pillai2024papagei}, or motif similarity \citep{xu2023rebar,xu2024relcon}. However, these require strong assumptions, either carefully designed augmentations or reliable positive selection strategies and are unable to do reconstruction out-of-box unlike the MAE methods.

\begin{figure}[!t]
    \centering
    \vspace{-5mm}
    \includegraphics[width=\textwidth]{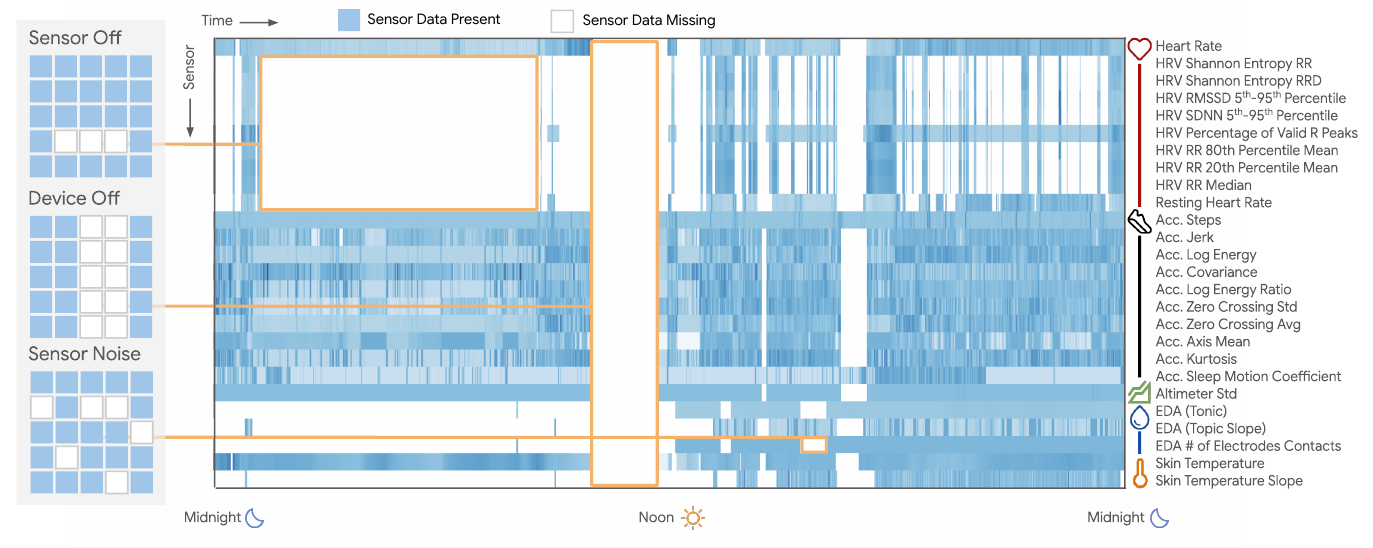}
    \vspace{-6mm}
    \caption{\textbf{The Fragmented Nature of Sensor Data.} Multimodal time-series sensor data frequently contains missing observations. Missingness can take several modes. In wearable data, these modes take the form of temporary periods in which a sensor(s) are off, periods in which the device is not warn, and measurements that are filtered out because they are clearly spurious/out of range.}
    \label{fig:data_examples} 
    \vspace{-3mm}
\end{figure}

\textbf{Learning from Incomplete Multimodal Data.} Our model learns general-purpose embeddings directly from incomplete multimodal time-series data through self-supervised pre-training, enabling effective transfer to diverse downstream tasks via simple linear probes. Existing representation learning works for incomplete data have focused primarily on either tabular data \citep{changlearning} or irregularly-sampled event time-series \citep{beebe2023paits}, both of which differ fundamentally from wearable sensors. Tabular missingness consists of simple, scattered, point-wise missingness, unlike the complex structured patterns in wearables, in which sensor groups across a time window will be missing and not at random (Figure~\ref{fig:data_examples}). While the irregularly-sampled domain shares some similarities, they have fundamentally different data characteristics. Irregularly sampled time-series such as ICU lab testing \citep{silva2012predicting} are collected at arbitrary intervals with all other modalities typically missing, whereas wearables produce regularly-sampled data where some modalities will drop out in structured groups (Figure~\ref{fig:data_examples}).

Alternatively, a separate body of incomplete multimodal data work has focused on learning imputation methods. The most relevant work is ReMasker \cite{du2023remasker}, which combines inherited and artificial masking in an MAE framework. Our approach differs in three fundamental aspects: (1) we optimize for representation learning rather than imputation, (2) we handle the complex missingness patterns characteristic of multimodal time-series (see Fig.~\ref{fig:data_examples}), as opposed to the simpler point-wise missingness in tabular data, and (3) we scale efficiently to long sequences (N=3744 tokens) compared to their limited context (N<20 tokens), representing a 35000x increase in compute (see Section \ref{sec:method} for details). Another approach, \citep{wei2024temporally}, similarly uses both inherited and artificial masks but limits attention to handcrafted time points (N=206) and uses self-attention blocks. While numerous deep learning methods exist for multivariate time-series imputation \citep{yoon2018gain, cao2018brits, qin2023imputegan, dai2024sadi}, these approaches focus solely on reconstruction quality and fail to produce general-purpose embeddings necessary for foundation models. \citep{jungo2024representation} investigate various imputation methods
and train classifiers on the reconstructed data, but do not learn representations for multiple downstream tasks. 
In contrast, our work handles real-world missingness patterns within a scalable representation learning framework.

\section{Large Scale Incomplete Wearable Data} \label{sec:datadescription}

\begin{wrapfigure}{b!}{0.37\textwidth}
    \vspace{-7mm}
    \centering
    \includegraphics[width=.35\textwidth]{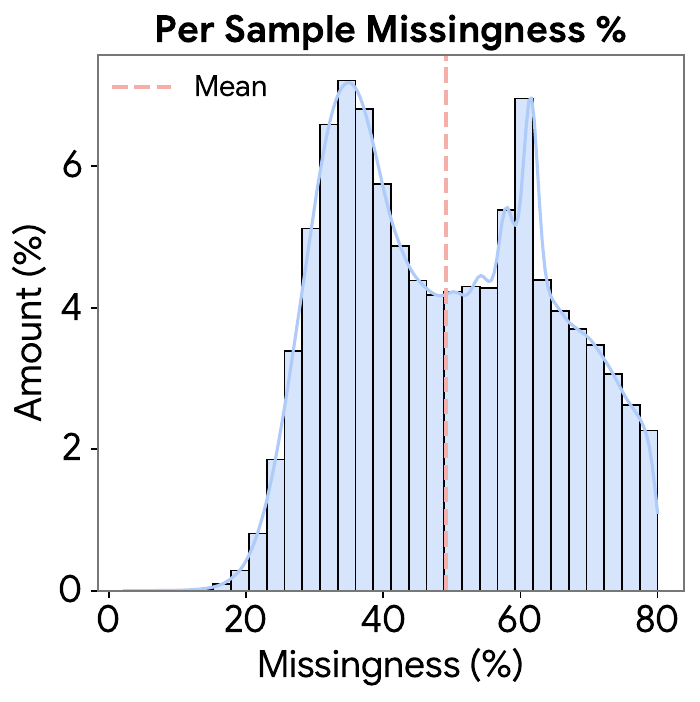}
    \caption{\textbf{Distribution of Missingness \% Per Sample.} Mean 49\%, Median 48\%, Std Dev 15\%, Minimum 2\%, Maximum 80\%. Samples with  $> 80\%$ missingness are discarded. }
    \vspace{-5pt}
    \label{fig:missperc} 
\end{wrapfigure}

\textbf{Data Summary.} A primary contribution of our work is in modeling incomplete data during pre-training, post-training, and inference. To validate our method we curate a large, unlabeled, pre-training dataset in addition to two labeled datasets for downstream tasks. Each data sample contains 26 minutely aggregated features from a set of 5 sensors (photoplethysmography, accelerometer, skin conductance, altimeter, and temperature) for a time span of 1440 minutes (1 day). A core property of these data is that they have complex, structured missingness patterns. A representative example of sensor data with missingness can be seen in Fig.~\ref{fig:data_examples}, along with the missingness distribution and statistics  shown in Fig. \ref{fig:missperc}. Missing data is ubiquitous in long-duration wearable sensor recordings, with 0\% of samples over our entire dataset of 1.6 million instances of 1 day data. All pre-trained and downstream datasets utilize similar devices and thus are subject to similar missingness patterns. Please refer to the Appendix for further data descriptions and statistics.

\textbf{Pre-training Data.} For pre-training, we used a de-identified dataset collected between March \nth{1} and May \nth{1} 2024 inclusive. The dataset included 3,581,748 person-days (or 40 million hours) sampled at minutely resolution from 60,440 people (37,352 men, 23,041 women, 47 unspecified). A mean of 59 days (min: 1, max: 93) were contributed per person with standard deviation of 32 days. All data used in this study were collected with the informed consent of research participants. This consent permits the use of data to generate findings for publication in scientific journals and other outlets, contributing to general knowledge about health and science. The mean reported participant age was 42.5 years (min: 18, max: 96 years, st.dev.: 12.6). The population reflects a wide range of body-mass index (BMI) values with 37\% healthy, 34\% overweight and 25\% obese in the training set and a similar cross-section in the validation set.

\textbf{Downstream Metabolic Study Data.} These data come from an IRB approved 
observational study of adults in the United States. We enrolled 4,416 participants, of which 1,250 had wearable data, labels and were included in our analysis. Demographics (age, BMI) and medical conditions (hypertension, anxiety) were collected via self-report.

\textbf{Downstream Activity Study Data.} These data come from the same source as our pretraining data. We randomly sampled approximately 5,000 examples for each of 20 activities for training and 1,000 examples of each activity for testing. The training and testing data were sampled in person-independent manner. The activities were from the following classes: \emph{Walking} \walkIcon, \emph{Bike} \bikeIcon, \emph{Playing Sports} \sportsIcon, \emph{Running} \runningIcon, \emph{Aerobics} \aerobicsIcon, \emph{Elliptical} \ellipticalIcon, \emph{Spinning} \spinningIcon, \emph{Weightlifting} \weightliftingIcon, \emph{Swimming} \swimIcon, \emph{Hiking} \hikingIcon, \emph{Playing Tennis} \tennisIcon, \emph{CrossFit} \crossfitIcon, \emph{Pilates} \pilatesIcon, \emph{Stairclimber} \stairclimberIcon, \emph{Dancing} \dancingIcon, \emph{Indoor climbing} \climbingIcon, \emph{Golf} \golfIcon, \emph{Skiing} \skiingIcon, \emph{Snowboarding} \snowboardingIcon, and \emph{Kayaking} \kayakingIcon.
In total, 104,086 activities were sampled from 46,199 people. The mean duration per activity was 66 minutes (min: 20 minutes, max: 360).

\section{Learning to \Largeprojectname with Adaptive Inherited Masking}
\label{sec:method}

\textbf{Motivation.} As sensor data frequently exhibits inherent missingness, our key idea is to inherit these missingness patterns to be used in conjunction with a masked pre-training framework~\citep{he2022masked}. These methods introduce an artificial mask on the present data and learn to reconstruct them. Artificial missingness sits in contrast to inherited missingness inherent to the data. Similar to the original MAE work \citep{he2022maskedmae}, our method implements an transformer-based encoder-decoder structure.

Our method first takes an input matrix of sensor features, which are then tokenized to be $\textbf{X} \in \mathbb{R}^{B \times N \times E}$ ($B$ is batch size, $N$ is number of tokens, and $E$ is embedding dimension). We then define a binary vector mask, $\textbf{M} \in \{0,1\}^{B \times N}$ (where 1 is masked and 0 is non-masked)
equal in length to the number of tokenized sensor inputs, where masked tokens are ignored by the encoder. Our method sets $\textbf{M}$ as the union of the inherited and artificial masks such that: 
\begin{equation*}
\textbf{M} = \textbf{M}^{\textrm{inherited}} \lor \textbf{M}^{\textrm{artificial}}  \label{eq:union}
\end{equation*}
The inherited mask, $\textbf{M}^{\textrm{inherited}}$, is the original, existing missingness present in the dataset. The artificial mask, $\textbf{M}^{\textrm{artificial}}$, is a simulated missingness on observed data. Critically, this inclusion of the inherited mask ensures that the encoder exclusively learns representations from reliable sensor data without contamination from imputation artifacts. 

\begin{wrapfigure}{r}{0.5\textwidth}
    \centering
    \vspace{-8mm}
    \captionof{table}{\textbf{Capabilities of Different Masking Implementations.} 
    We combine dropout removal's efficiency \citep{he2022masked} with attention masking's flexibility \citep{du2023remasker} to allow us to process to long sequences with inherited masks that have varying mask \%. 
    }
    \resizebox{.5\textwidth}{!}{%
    \begin{tabular}{rcccc}
    \toprule[1.5pt]
        \textbf{Masking} & {\textbf{\shortstack{Complexity}}} & {\textbf{\shortstack{Fixed\\Mask \%}}} & {\textbf{\shortstack{Dynamic\\Mask \%}}} & {\textbf{\shortstack{Allows\\Inherited\\Mask}}} \\
        \midrule    
        Dropout Removal \citep{he2022masked}& \textcolor{black}{$O((N-D)^2)$} & \cmark & \xmark & \xmark \\         \addlinespace[2pt] 
        Attention Mask \citep{du2023remasker}& \textcolor{black}{$O(N^2)$} & \cmark & \cmark & \cmark \\         \addlinespace[2pt] 
        \grayrow
         \textbf{\projectname (ours)} & \textcolor{black}{$O((N-D)^2)$} & {\textbf{\cmark}}  & {\textbf{\cmark}}  & {\textbf{\cmark}}  \\
    \bottomrule[1.5pt]
    \end{tabular}%
    }
    \label{tab:computational_efficiency}
    \footnotesize \vspace{1.5pt}
    \newline
    \footnotesize
    $N$: Number of tokens $D$: Number dropped
    \vspace{-2mm}
\end{wrapfigure}

\begin{figure*}[!t]
    \centering
    \vspace{-5mm}
    \includegraphics[width=\textwidth]{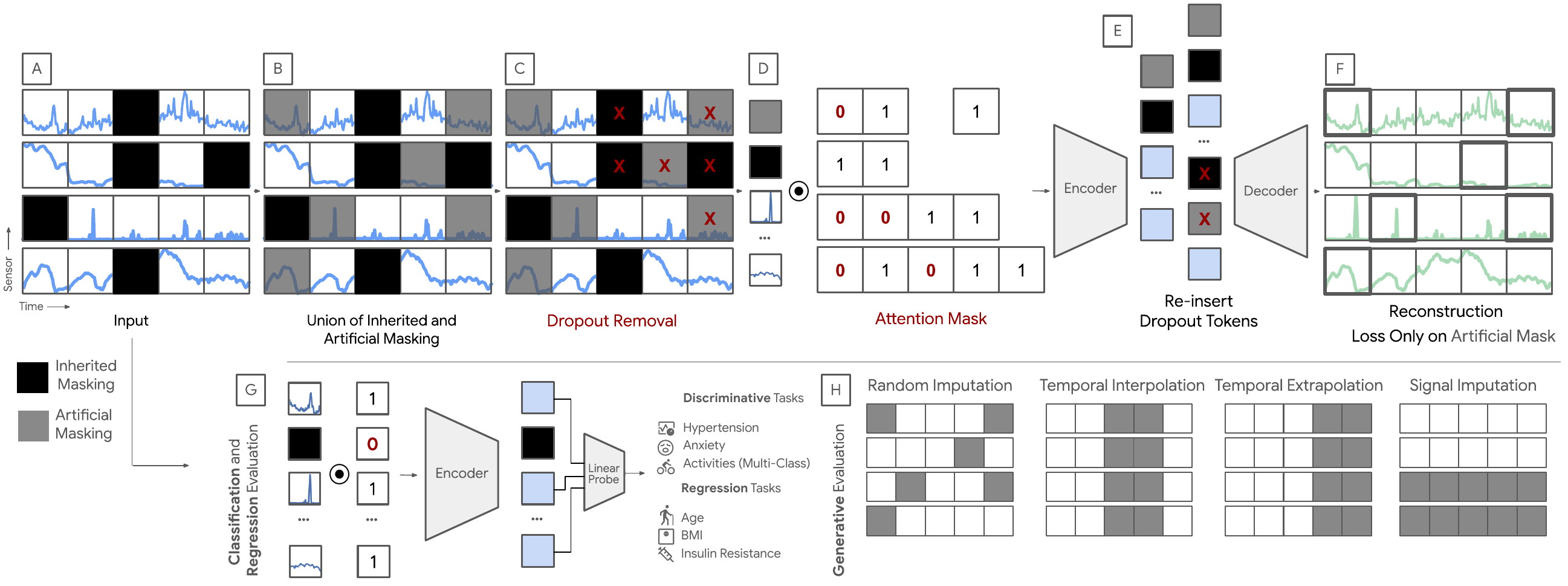}
    \caption{\textbf{LSM-2 Pre-training with \projectname [A-F] and Evaluation [G,H]}. Our mask is a union of \textbf{[A]} inherited missingness from real-world noise  and \textbf{[B]} artificial masking of observed data. Both are modeled with identical, learnable tokens. Because the inherited mask introduces variable masking, \textbf{[C]} we first remove $D$ (size of artificial mask) tokens  and \textbf{[D]} then use an attention mask to remove the remaining. \textbf{[E]}  Dropped tokens are reinserted before \textbf{[F]} the final reconstruction. \textbf{[G]} Reconstruction error is computed only on artificial masks with known ground truth.
    \textbf{[H]} For predictive evaluations, a linear probe is trained on a pooled representation of the non-missing data. 
    \vspace{-3pt} 
    }
    \label{fig:architecture} 
\end{figure*}

\textbf{Background.} The original MAE work~\cite{he2022masked} implements masking through \emph{dropout removal},
where masked tokens are not passed through the encoder. Specifically it assumes that a fixed number of tokens $D$ are dropped for every sample, such that $\sum_{n=1}^{N} \textbf{M}_{[b,n]} = D\ \forall\ b \in [1, B]$. The transformer encoder input can then be formulated as $\textbf{X}_{[\textbf{M}, :]} \in \mathbb{R}^{B \times (N - D) \times E}$. This reduces the transformer's computational complexity from $O(N^2) \rightarrow O((N-D)^2)$, which translates to 25x less computation when masking 80\% of tokens ($D=0.8N$). While efficient, this approach requires fixed masking amount $D$, in order to construct batched encoder input $\textbf{X}_{[\textbf{M}, :]}$ with $B>1$.  The motivation of our \projectname approach is to include inherited masking in the MAE procedure in order to model real-world missingness. However, we are unable to do so with dropout removal because the amount of pre-existing missingness will vary, and causing the $D$ of the inherited mask also vary.
Recent methods have attempted to handle variable masking~\cite{du2023remasker} by utilizing the transformer's \emph{attention masking} mechanism \citep{vaswani2017attention}. While flexible, these methods fail to use dropout removal, making them computationally prohibitive for long sequences and large scale pre-training.

\textbf{Adaptive Attentive Masking Design.} Our key insight is to combine both masking modes in a unified approach: we maintain dropout removal's efficiency while incorporating the flexibility of attention masking. This hybrid strategy is visualized in Figure~\ref{fig:architecture}. Dropout removal limits the number of tokens that must be encoded to the lower bound of artificially masked tokens.
This is because the set of dropped tokens $D$ is static. In scenarios where a sample has no inherent missing data, these dropped tokens must be entirely defined by the artificial mask.
In practice, dropped-out tokens can be a mix of inherited and artificially masked tokens. Similarly, the remaining masked tokens, which are disregarded using the transformer's attention mask, can also be of either type. This fusion provides the benefits of both paradigms while mitigating their individual limitations.

\textbf{Unified Framework for Pre-training and Evaluation.}
\projectname provides a unified framework for LSM-2 that consistently handles missing data during both pre-training and inference. The full pre-training procedure can be seen in Figure \ref{fig:architecture} [A-G]. During pre-training, the adaptive masking not only enables the inclusion of varying inherited mask sizes, but also allows the artificial masking to include a mix of strategies with differing masking percentages. Our artificially masking mix seeks to model the real-world missingness patterns shown in Figure \ref{fig:data_examples}. The mix includes (1) 80\% random imputation masking (to model noise), where a random patch is masked, (2) 50\% temporal slice masking (to model off body), where all sensors at a random time point are masked, and (3) 50\% signal slice masking (to model sensor off), where all time points for a random sensor are masked. Each instance uses a randomly selected masking strategy with equal probability. The specific masking percentages were identified via an ablation study, reported within the Appendix. As such, we set $D=0.5N$, boosting our computational efficiency by 4x.

Crucially, \projectname's adaptive masking is also used during evaluation, which can be seen in Figure \ref{fig:architecture} [G,H]. The pre-trained model is then able to operate directly on incomplete multimodal sensor data by dynamically attending only to observed segments. This eliminates the need for external preprocessing, such as imputing or discarding missing values, and ensures generalization from pre-training to downstream deployment in real-world settings.



\section{Experiments}
\label{sec:Experiments}

\textbf{Pre-training Set-up.} We pre-train LSM-2 on minutely multimodal wearable data (\textbf{A} $\in \mathbb{R}^{T \times S}$) where $S=26$ sensor features and $T=1440$ minutes. Inputs are tokenized using a ViT-1D \citep{dosovitskiy2020image, abbaspourazad2023large} encoder with a 1D patch size of 10 minutes, resulting in 3744 tokens (144 tokens per signal). We apply a shared kernel across channels and use a 2D positional embedding to encode time and signal identity. The model has 25M parameters, 384-d hidden size, 12 encoder layers, and 4 decoder layers. Following Section~\ref{sec:method}, we apply a composite mask (80\% random, 50\% temporal, 50\% signal slices) and optimize mean squared error over masked patches on reconstruction. Notably, we do not back-propagate on missing pixels for any of the SSL methods trained including baselines. Training is performed on 8x16 Google v5e TPUs with a batch size of 512 for 100K steps. SSL baselines—LSM-1 \citep{narayanswamy2024scaling}, SimCLR \citep{chen2020simple}, DINO \citep{caron2021emerging}, and MSN \citep{assran2022masked}—are trained from scratch using the same setup unless otherwise noted. LSM-1 uses a ViT-2D with a (10,2) patch size and 0.8 random masking, while the contrastive methods rely on jittering, scaling, and time-flipping augmentations \citep{tang2020exploring, liu2024guidelines, zhang2022self, rommel2022data}. All baselines use imputed data to meet their full-input requirement. See Appendix for further implementation details.

\textbf{Downstream Evaluation.} We evaluate LSM-2 across three downstream targets: generative, classification, and regression. For \texttt{generative}, we assess reconstruction under structured missingness patterns: (1) random imputation (30\%, 50\%, 80\%), (2) temporal interpolation (contiguous masked windows of 10, 30, or 60 minutes), (3) temporal extrapolation (masked window at the end of the sequence), and (4) signal imputation (masking 2/26, 6/26, or 12/26 channels). Since contrastive baselines lack reconstruction objectives, we compare against LSM-1~\citep{narayanswamy2024scaling} in addition to simple imputation methods used in practice—Linear Interpolation, Nearest Neighbors, and Mean Filling—under the same union masking scheme. We omit MICE~\citep{van2011mice} due to its missingness at random assumptions not holding and its lower performance in prior work~\citep{narayanswamy2024scaling}. For \texttt{classification}, we average embeddings over non-inherited-masked tokens and apply a trainable linear probe; LSM-1 pools across all tokens, and contrastive methods use the CLS token. We report F$_1$, Accuracy, Balanced Accuracy, and AUROC on targets including hypertension, anxiety (Metabolics dataset; see Section~\ref{sec:datadescription}), and 20-class activity recognition (Activity dataset). For \texttt{regression}, we follow the same setup with a linear regression probe and report MAE and Pearson correlation on BMI and age (Metabolics dataset). See Appendix for further details.

\section{Results and Discussion} \label{sec:resultsanddiscussion}

\textbf{Generalizability across classification, generative, and regression tasks.} LSM-2 with \projectname learns a strong generalizable representation, useful for classification, regression and generative tasks (Tables ~\ref{tab:classification_tasks}, \ref{tab:generative_tasks_compact}, \ref{tab:regression_selected} respectively).  This research presents preliminary findings and should not be interpreted as providing diagnostic tools or recommendations.

Due to our improved pre-training reconstruction objective, LSM-2 obtains much stronger generative results compared to the prior state-of-the-art work - LSM-1 \cite{narayanswamy2024scaling} which was limited in its masking strategy (artificial random imputation masking). By introducing a mixture of artificial masking strategies with flexible missing ratios, as well as the inclusion of the inherited mask, not only do we achieve a \textbf{+33\% performance increase} on the 80\% random imputation evaluation, but we also achieve strong benefits across different generative tasks, with \textbf{+77\% improvement} in 2 signal imputation and a  \textbf{+47\% improvement} in 10 minute temporal interpolation. This demonstrates that explicitly modeling diverse missingness patterns during pre-training leads to more robust representations that generalize better to real-world scenarios with complex data gaps.

\begin{table}[t!]
\centering
\vspace{-1mm}
\setlength{\tabcolsep}{3pt}
\captionof{table}{\textbf{Classification Task Results}}
\vspace{-2mm}
\label{tab:classification_tasks}
\small
\resizebox{\textwidth}{!}{
\begin{tabular}{c lcccc|cccc|cccc}
\toprule[1.5pt]
& & \multicolumn{4}{c|}{Hypertension (2)} & \multicolumn{4}{c|}{Anxiety (2)} & \multicolumn{4}{c}{Activity Recognition (20)} \\
\cmidrule(lr){3-6} \cmidrule(lr){7-10} \cmidrule(lr){11-14}
& Method & \contour{black}{$^{\uparrow}$}F$_1$ & \contour{black}{$^{\uparrow}$}Acc & \contour{black}{$^{\uparrow}$}BAcc & \contour{black}{$^{\uparrow}$}AUC & \contour{black}{$^{\uparrow}$}F$_1$ & \contour{black}{$^{\uparrow}$}Acc & \contour{black}{$^{\uparrow}$}BAcc & \contour{black}{$^{\uparrow}$}AUC & \contour{black}{$^{\uparrow}$}F$_1$ & \contour{black}{$^{\uparrow}$}Acc & \contour{black}{$^{\uparrow}$}BAcc & \contour{black}{$^{\uparrow}$}AUC \\
\midrule\midrule
\multirow{2}{*}{\rotatebox[origin=c]{90}{\textbf{ST}}}
& ResNet & 0.516 & 0.529 & 0.587 & 0.624 & 0.645 & 0.655 & 0.651 & 0.709 & \textbf{0.729} & \textbf{0.721} & \textbf{0.734} & \textbf{0.965} \\
& ViT1D & 0.481 & 0.516 & 0.509 & 0.520 & 0.583 & 0.597 & 0.586 & 0.620 & 0.351 & 0.367 & 0.374 & 0.863 \\
\midrule
\multirow{5}{*}{\rotatebox[origin=c]{90}{\textbf{LP}}}
& SimCLR & 0.501 & 0.524 & 0.548 & 0.568 & 0.594 & 0.603 & 0.601 & 0.636 & 0.098 & 0.109 & 0.124 & 0.603 \\
& DINO & 0.487 & 0.536 & 0.504 & 0.510 & 0.551 & 0.557 & 0.562 & 0.582 & 0.102 & 0.110 & 0.124 & 0.635 \\
& MSN & 0.512 & 0.553 & 0.538 & 0.552 & 0.579 & 0.585 & 0.588 & 0.622 & 0.108 & 0.118 & 0.125 & 0.662 \\
& \textsc{LSM-1} & 0.640 & 0.676 & 0.682 & 0.739 & 0.670 & 0.678 & 0.678 & 0.743 & 0.470 & 0.470 & 0.489 & \textbf{0.900} \\
\grayrow
& \textsc{LSM-2} & \textbf{0.651} & \textbf{0.687} & \textbf{0.693} & \textbf{0.754} & \textbf{0.683} & \textbf{0.690} & \textbf{0.692} & \textbf{0.758} & \textbf{0.474} & \textbf{0.472} & \textbf{0.493} & 0.899 \\
\midrule
& $\Delta$ LSM-1 & 
\textcolor{textgreen}{\textbf{+1.7\%}} & \textcolor{textgreen}{\textbf{+1.6\%}} & \textcolor{textgreen}{\textbf{+1.6\%}} & \textcolor{textgreen}{\textbf{+2.0\%}} &
\textcolor{textgreen}{\textbf{+1.9\%}} & \textcolor{textgreen}{\textbf{+1.8\%}} & \textcolor{textgreen}{\textbf{+2.1\%}} & \textcolor{textgreen}{\textbf{+2.0\%}} &
\textcolor{textgreen}{\textbf{+0.8\%}} & \textcolor{textgreen}{\textbf{+0.4\%}} & \textcolor{textgreen}{\textbf{+0.8\%}} & \textcolor{textred}{\textbf{-0.1\%}} \\
\bottomrule[1.5pt]
\end{tabular}} \label{tbl:class}
\fontsize{7}{7}\selectfont 
\vspace{3pt}

Metrics: F$_1$ Score, Accuracy, Balanced Accuracy, AUROC with Macro One-vs-Rest $\vert$ Tasks: 20-class Activity Recognition, rest are binary $\vert$ Methods: Supervised Training (ST), Linear Probe (LP).
\vspace{0mm}
\end{table}

\begin{table}[t]
\centering
\centering
\setlength{\tabcolsep}{1pt}
    \vspace{0mm}
\captionof{table}{\textbf{Generative Task Results}}
\vspace{-2mm}
\label{tab:generative_tasks_compact}
\small
\resizebox{\linewidth}{!}{%
\begin{tabular}{lccc|ccc|ccc|ccc}
\toprule[1.5pt]
& \multicolumn{3}{c|}{\contour{black}{$^{\downarrow}$}Random Imp.} & \multicolumn{3}{c|}{\contour{black}{$^{\downarrow}$}Temporal Interp.} & \multicolumn{3}{c|}{\contour{black}{$^{\downarrow}$}Temporal Extrap.} & \multicolumn{3}{c}{\contour{black}{$^{\downarrow}$}Signal Imp.} \\
\cmidrule(lr){2-4} \cmidrule(lr){5-7} \cmidrule(lr){8-10} \cmidrule(lr){11-13}
Method & 30\% & 50\% & 80\% & 10m & 30m & 60m & 10m & 30m & 60m & 2 & 6 & 12 \\
\midrule\midrule
Linear Int. & 0.57 & 0.62 & 0.74 & 0.42 & 0.56 & 0.70 & 0.47 & 0.64 & 0.82 & NA & NA & NA \\
NN Fill & 0.70 & 0.76 & 0.90 & 0.52 & 0.69 & 0.84 & 0.47 & 0.64 & 0.82 & NA & NA & NA \\
Mean Fill & 0.92 & 0.96 & 0.93 & 0.79 & 0.80 & 0.84 & 0.78 & 0.80 & 0.83 & 1.28 & 1.30 & 1.29 \\
\midrule
LSM-1 & 0.21 & 0.24 & 0.30 & 0.49 & 0.55 & 0.60 & 0.45 & 0.52 & 0.56 & 0.73 & 0.58 & 0.45 \\
\grayrow
\textsc{LSM-2} & \textbf{0.18} & \textbf{0.20} & \textbf{0.20} & \textbf{0.26} & \textbf{0.37} & \textbf{0.45} & \textbf{0.28} & \textbf{0.38} & \textbf{0.48} & \textbf{0.17} & \textbf{0.21} & \textbf{0.27} \\
\midrule
$\Delta$ LSM-1 &
\textcolor{textgreen}{\textbf{+14\%}} & \textcolor{textgreen}{\textbf{+17\%}} & \textcolor{textgreen}{\textbf{+33\%}} &
\textcolor{textgreen}{\textbf{+47\%}} & \textcolor{textgreen}{\textbf{+31\%}} & \textcolor{textgreen}{\textbf{+25\%}} &
\textcolor{textgreen}{\textbf{+38\%}} & \textcolor{textgreen}{\textbf{+27\%}} & \textcolor{textgreen}{\textbf{+14\%}} &
\textcolor{textgreen}{\textbf{+77\%}} & \textcolor{textgreen}{\textbf{+64\%}} & \textcolor{textgreen}{\textbf{+40\%}} \\
\bottomrule[1.5pt]
\end{tabular}%
}
\fontsize{6}{6}\selectfont 
\vspace{3pt}

Metrics: Mean Squared Error $\vert$ Tasks: Random Imputation (30\%, 50\%, 80\% missing), Temporal Interpolation/Extrapolation (10, 30, 60 missing minutes), Signal Imputation (2, 6, or 12 out of 26 missing modalities) $\vert$ Methods: Linear interpolation, Nearest neighbor fill, Mean filling
\vspace{-8pt}

\end{table}

\begin{wrapfigure}{r}{0.45\textwidth} 
    \centering
    \vspace{-0mm}
    \captionof{table}{\textbf{Regression Task Results}}
    \vspace{-2mm}
    \label{tab:regression_selected}
    \small
    \resizebox{\linewidth}{!}{%
    \begin{tabular}{clcc|cc}
    \toprule[1.5pt]
    & & \multicolumn{2}{c|}{Age} & \multicolumn{2}{c}{BMI} \\
    \cmidrule(lr){3-4} \cmidrule(lr){5-6}
    & Method & \contour{black}{$^{\downarrow}$}MAE & \contour{black}{$^{\uparrow}$}Corr & \contour{black}{$^{\downarrow}$}MAE & \contour{black}{$^{\uparrow}$}Corr \\
    \midrule\midrule
    \multirow{2}{*}{\rotatebox[origin=c]{90}{\textbf{ST}}}
    & ResNet & 7.43 & 0.618 & 5.07 & 0.515 \\
    & ViT1D & 9.65 & 0.132 & 6.06 & 0.047 \\
    \midrule
    \multirow{5}{*}{\rotatebox[origin=c]{90}{\textbf{LP}}}
    & SimCLR & 9.21 & 0.345 & 5.85 & 0.235 \\
    & DINO & 9.69 & 0.112 & 5.97 & 0.122 \\
    & MSN & 9.42 & 0.255 & 5.84 & 0.250 \\
    & \textsc{LSM-1} & \textbf{6.41} & \textbf{0.728} & 4.39 & 0.667 \\
    \grayrow
    & \textsc{LSM-2} & 6.49 & 0.722 & \textbf{4.38} & \textbf{0.673} \\
    \midrule
    & $\Delta$ LSM-1 & 
    \textcolor{textred}{\textbf{-1.2\%}} & \textcolor{textred}{\textbf{-0.8\%}} &
    \textcolor{textgreen}{\textbf{+0.2\%}} & \textcolor{textgreen}{\textbf{+1.0\%}} \\
    \bottomrule[1.5pt]
    \end{tabular}%
    }
    \fontsize{6}{6}\selectfont 
    Metrics: Mean Absolute Error, Pearson Correlation $\vert$ Methods: Supervised Training (ST), Linear Probe (LP).
    \vspace{-10pt}
\end{wrapfigure}

Despite being pre-trained on with a reconstruction objective, LSM-2 achieves SOTA performance across classification tasks, \textbf{beating all other self-supervised learning baselines}. Even with a simple linear probe and frozen features, our model surpasses fully supervised baselines on hypertension and anxiety prediction — two challenging tasks that previously required hand-crafted features or custom architectures \citep{silva2022machine, abd2023wearable}. This suggests that pre-training helps avoid overfitting and enables the model to capture subtle physiological cues that generalize across conditions. The strong results across both binary (hypertension/anxiety) and multi-class (activity recognition) tasks indicate that the model learns hierarchical features suited to different levels of task complexity.

In regression tasks, LSM-2 improves correlation on BMI by +1.0\%, while underperforming on age prediction by -0.8\%. Since the absolute metric (e.g., mean absolute error) is affected by differing target scales (e.g., Age: 18--90 vs. BMI: 12--65), correlation offers a clearer view of model quality. 

\begin{figure*}[t!]
    \centering
    \vspace{-5mm}
    \includegraphics[width=\textwidth]{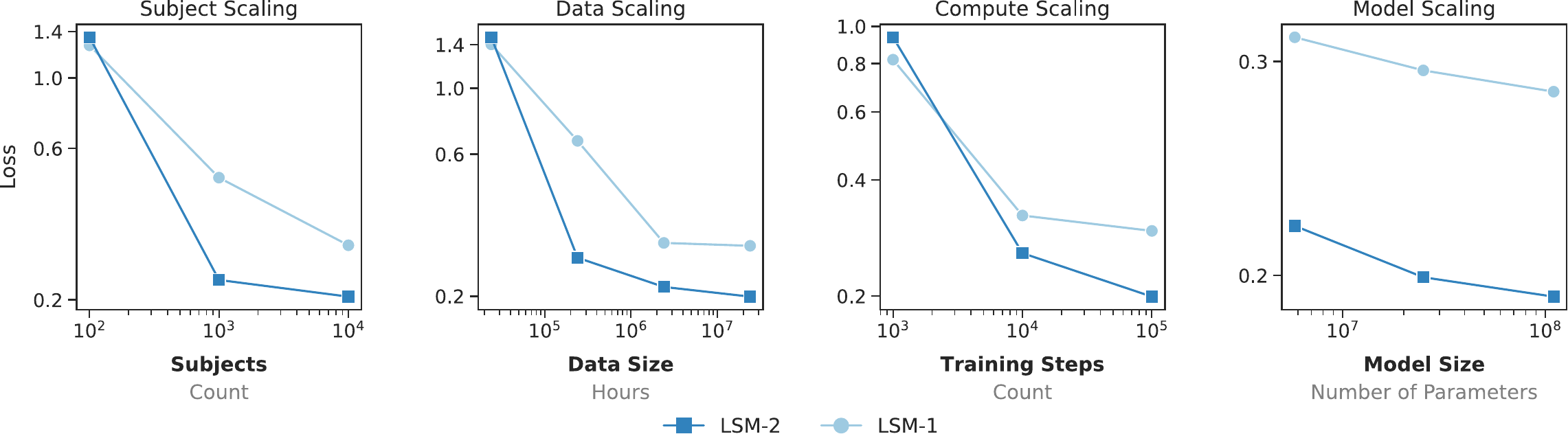}
    \caption{\textbf{Scaling Performance of Our Model.} LSM-2 model achieves better scaling than LSM-1 across all dimensions: \textit{subjects}, \textit{data}, \textit{compute}, and \textit{model size}. LSM-2 uses a mixed masking strategy during  pre-training, but here we report only random imputation loss to match LSM-1. 
    }
    \label{fig:scaling_results} 
    \vspace{-5mm}
\end{figure*}

\textbf{Strong scaling performance on 40 million hours of incomplete data.} Figure \ref{fig:scaling_results} show that our \projectname scales more effectively than the LSM-1 model across 4 different dimensions: subject, data, compute, and model. The LSM-1 model exhibits scaling saturation for the data and compute dimensions, but our model's trend indicates a more aggressive downwards slope that has not yet saturated. These results are promising as they suggest that continued investment in larger datasets and compute may yield further performance gains, indicating that our method has not yet reached its  limits.

\begin{figure*}[!t]
    \centering
    \vspace{-2mm}
    \includegraphics[width=.95\textwidth]{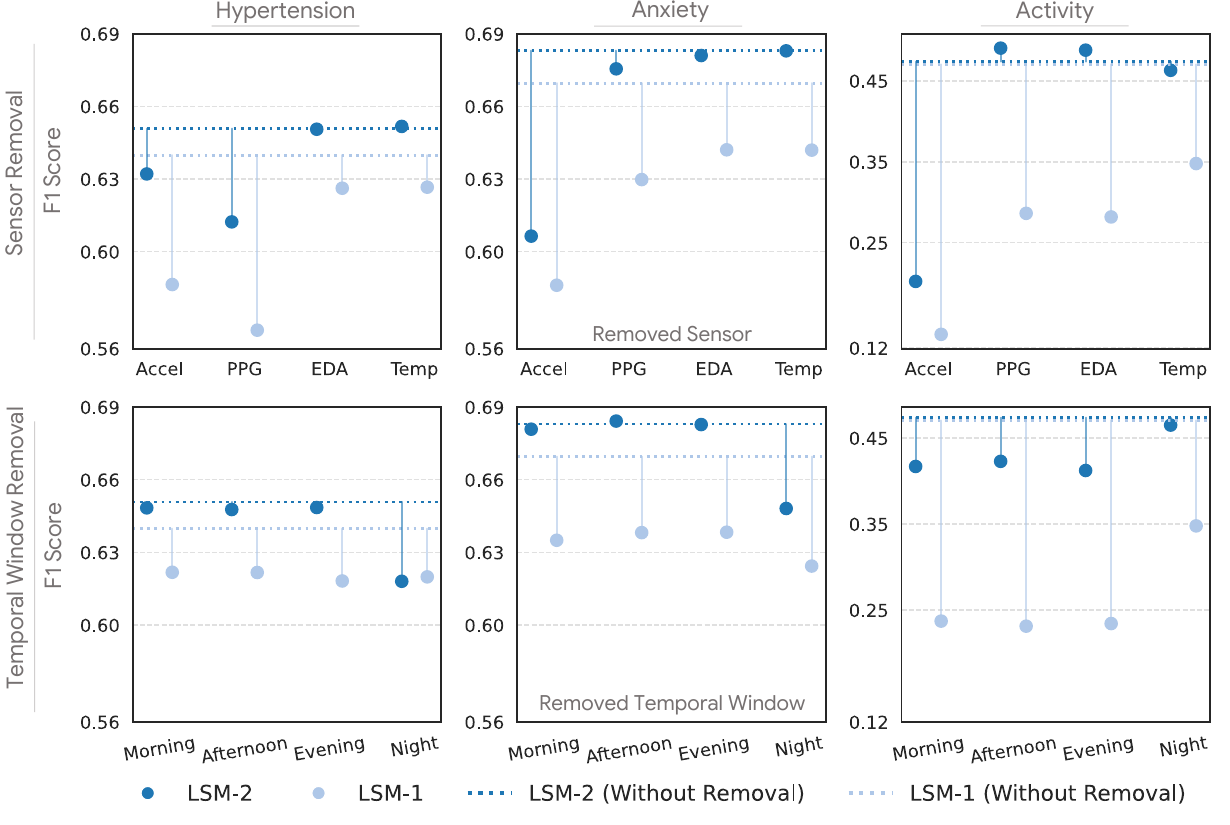}
    \caption{\textbf{Robustness to Targeted Missingness.} In sensor removal, all signals derived from the specific sensor are removed. In temporal window removal, all signals are removed at a given timeframe (Morning [8am-12pm], Afternoon [12pm-4pm], Evening [4pm-8pm], Night [8pm-8am]). The dotted line denotes a model trained on all modalities. When evaluating with simulated sensor- or time-specific missingness, LSM-2 maintains consistent performance while LSM-1 degrades significantly. Where LSM-2 does show sensitivity, it aligns with domain knowledge. For example, nighttime BP's stronger predictive power of hypertension over daytime \citep{hansen2011predictive}, accelerometry's role in distinguishing anxiety from physiological stress responses \citep{sevil2020detection}.}
    \label{fig:mnar} 
    \vspace{-2mm}
\end{figure*}

\textbf{Strong Robustness to Targeted Missingness.} LSM-2 with \projectname demonstrates substantially greater resilience to targeted missingness compared to prior work, as seen in Figure \ref{fig:mnar}. Across 11 out of 12 missingness scenarios, our model consistently maintains stronger performance. For example, when accelerometry is removed—a key sensor for activity recognition—our model's F$_1$ score drops from 0.47 to 0.20 (\textminus57\%), while LSM-1 degrades more severely from 0.47 to 0.14 (\textminus71\%). Notably, even in this degraded setting, our model still outperforms LSM-1 by \textbf{+47\%} in absolute terms. A similar trend holds across other modalities: removing PPG during hypertension prediction leads to only a \textminus6\% drop for \projectname (0.65 to 0.61), compared to \textminus11\% for LSM-1 (0.64 to 0.57).

Robustness also generalizes across temporal ablations. While both models reach similar peak activity recognition scores ($\sim$0.47 F$_1$), our model maintains an average F$_1$ of 0.43 across temporal ablations—substantially higher than LSM-1's 0.26 (\textbf{+65\%} relative gain). Overall, these results validate the effectiveness of our adaptive masking strategy in modeling missingness patterns. Our model experiences \textbf{73\% smaller performance drops} across all 12 ablation settings and retains \textbf{+15\% higher} absolute performance in degraded states. This combination of robustness and accuracy makes \projectname a more reliable choice for real-world deployment, where missing data is a reality.

\textbf{Reflects Physiological Domain Knowledge and Other Real-world Implications.} The targeted missingness experiments in Figure \ref{fig:mnar} also reveal clinically coherent patterns with real-world implications. LSM-2's hypertension and anxiety predictions show the expected nocturnal advantage, such that the removal of nighttime signals has 5\% degradation in F$_1$ for both targets, compared to an average 0.4\% and 0.01\% degradation for the daytime windows for each target. This finding strongly aligns with clinical literature demonstrating the diagnostic value of nighttime biosignals for  hypertension \citep{yilmaz2023nocturnal, hansen2011predictive} and stress prediction \citep{kinnunen2020feasible, fan2024sleep}, which are less affected by daily activity artifacts and better capture underlying pathophysiology.

Interestingly, LSM-2 also demonstrates a large 11\% drop in performance for anxiety prediction after removing the accelerometry sensor, whereas removing the other sensors only results in an average 0.5\% drop. This suggests accelerometry provides unique signals for anxiety detection that are not captured by other modalities. There have been recent research works \citep{sevil2020detection, wu2015modeling} that demonstrate the importance of utilizing accelerometry sensors in stress prediction in order to distinguish anxiety and mental stress from physiological stress responses from physical activity.

These results demonstrate three key advantages of our \projectname adaptive masking approach: (1) performance degrades proportionally to a sensor's clinical importance, (2) cross-modal relationships are maintained when inputs are missing, and (3) known temporal biases in physiological data are preserved. This robustness is crucial for real-world deployment where missing data is inevitable, making \projectname significantly more reliable in field settings.

\begin{wrapfigure}{r}{0.45\textwidth}
    \centering
    \vspace{-15pt}
    \captionof{table}{\textbf{Ablation Study}}
    \vspace{-7pt}
    \label{tab:ablation_study}
    \small
    \setlength{\tabcolsep}{4pt}
    \renewcommand{\arraystretch}{1.15}
    \resizebox{\linewidth}{!}{%
    \begin{tabular}{@{}l *{4}{>{\centering\arraybackslash}m{1.62cm}@{}}}
    \toprule[1.5pt]
    & \multicolumn{2}{c}{\textbf{Generative (\contour{black}{$^{\downarrow}$}MSE})} & \multicolumn{2}{c}{\textbf{Classification (\contour{black}{$^{\uparrow}$}F$_1$)}} \\
    \cmidrule(lr){2-3} \cmidrule(lr){4-5}
    & \begin{tabular}{@{}c@{}}80\% \\ R.Imp.\end{tabular} & \begin{tabular}{@{}c@{}}60m \\ T.Interp.\end{tabular} & Anxiety & Activity \\
    \midrule
    \rowcolor[gray]{0.9}
    \textsc{\projectname} & 0.20 & 0.45 & 0.683 & 0.474 \\
    \midrule
    w/o Inheritance & 
    \textcolor{textred}{0.28} & \textcolor{textred}{0.62} & 
    \textcolor{textred}{0.671} & \textcolor{textred}{0.445} \\
    w/o Mixing & 
    \textcolor{textgreen}{0.19} & \textcolor{textred}{0.58} & 
    \textcolor{textred}{0.637} & \textcolor{textred}{0.460} \\
    \bottomrule[1.5pt]
    \end{tabular}%
    }
    \vspace{-20pt}
\end{wrapfigure}
\textbf{Importance of Inheritance and Mask Mixing. }\projectname is composed of two main components: (1) inclusion of an Inherited Mask and (2) usage of a mix of artificial masking with randomly using either 80\% random imputation, 50\% temporal slices, or 50\% signals slices. In Table \ref{tab:ablation_study}, we show how removing inheritance leads to performance degradation across all of the various tasks. Without mixing, only an 80\% random imputation pre-training task is used, matching prior work\citep{narayanswamy2024scaling}. While the random imputation performance improves, all other tasks degrade, including the other generative task, temporal interpolation.

\textbf{Limitations and Future Work.} Our study has several important constraints. First, training and evaluation were limited to a specific private datasets, necessitating future work on exploring other datasets with complex missingness patterns, such as All of Us \citep{Jeong2025allofus}, and understanding missingness distribution shifts. Furthermore, we make use of minutely aggregated features, which is helpful for helping us model our 1-day longer time-scale day data, but uncommon in the broader wearable sensing space, which focuses primarily on raw high frequency sensor signal. Unfortunately, this is a practical limitation, as data is not stored in its raw form at such scale. Finally, although the focus of our work is on multimodal sensor data, our technique is broadly applicable and domain-agnostic requiring only that the data contains existing missingness. Therefore, future work can explore the application of our \projectname across different missingness-afflicted domains.

\section{Conclusion}

In this work, we introduced the second generation of Large Sensor Model (LSM-2) with \underline{A}daptive and \underline{I}nherited \underline{M}asking, {\large\texttt{AIM}}, a novel self-supervised learning approach designed to learn robust representations directly from incomplete wearable sensor data. By integrating both inherited (real-world) and artificial masking strategies, AIM eliminates the need for explicit imputation while effectively modeling the pervasive missingness in real-world sensor data. Our experiments demonstrate that our foundation model LSM-2, pre-trained with \projectname, achieves state-of-the-art performance and scaling capability across a diverse range of tasks across differing semantics. Our targeted missingness experiments reveal that LSM-2 maintains strong performance even when entire sensors are dropped, suggesting broad applicability to scenarios with varying sensor availability. Our model's strong performance under real-world missingness conditions demonstrates its practical applicability, and we hope the insights in our work will guide future work in machine learning methodologies for wearable sensors and health time-series. 

\bibliographystyle{abbrv}
\bibliography{references}
\newpage

\resumetoc

\setlength{\cftsecnumwidth}{4.5ex}
\setlength{\cftsubsecnumwidth}{6ex}

\newcommand{\appendixheader}{%
  \vbox{%
    \hsize\textwidth
    \linewidth\hsize
    \vskip 0.1in
    \hrule height 4pt 
    \vskip 0.25in
    \centering
    {\LARGE\bfseries Appendix --- LSM-2: Learning from Incomplete Sensor Data\par} 
    \vskip 0.29in
    \hrule height 1pt 
    \vskip 0.09in
  }%
}
\appendixheader 

\renewcommand{\contentsname}{Table of Contents}
{
\hypersetup{hidelinks}
\tableofcontents
}

\clearpage

\appendix

\renewcommand{\thesection}{A.\arabic{section}}

\makeatletter

\section{Data Details}

\subsection{Imputing Missingness for Non \projectname Models}

Although \projectname is able to organically handle existing missing values using clever masking, the same cannot be said for our baseline methods. Furthermore, many standard deep learning frameworks (such as pytorch, jax, and tensorflow) are unable to handle nan values in model training and evaluation, causing value errors or propogating nans throughout the network during forward and backward passes. For this reason we impute missing (nan) values in our data. We use linear interpolation between gaps and then back and forward fill for missingness at the start and end of the sequence.

\subsection{Device Details}

\begin{figure*}[!htbp]
    \centering
    \includegraphics[width=0.5\textwidth]{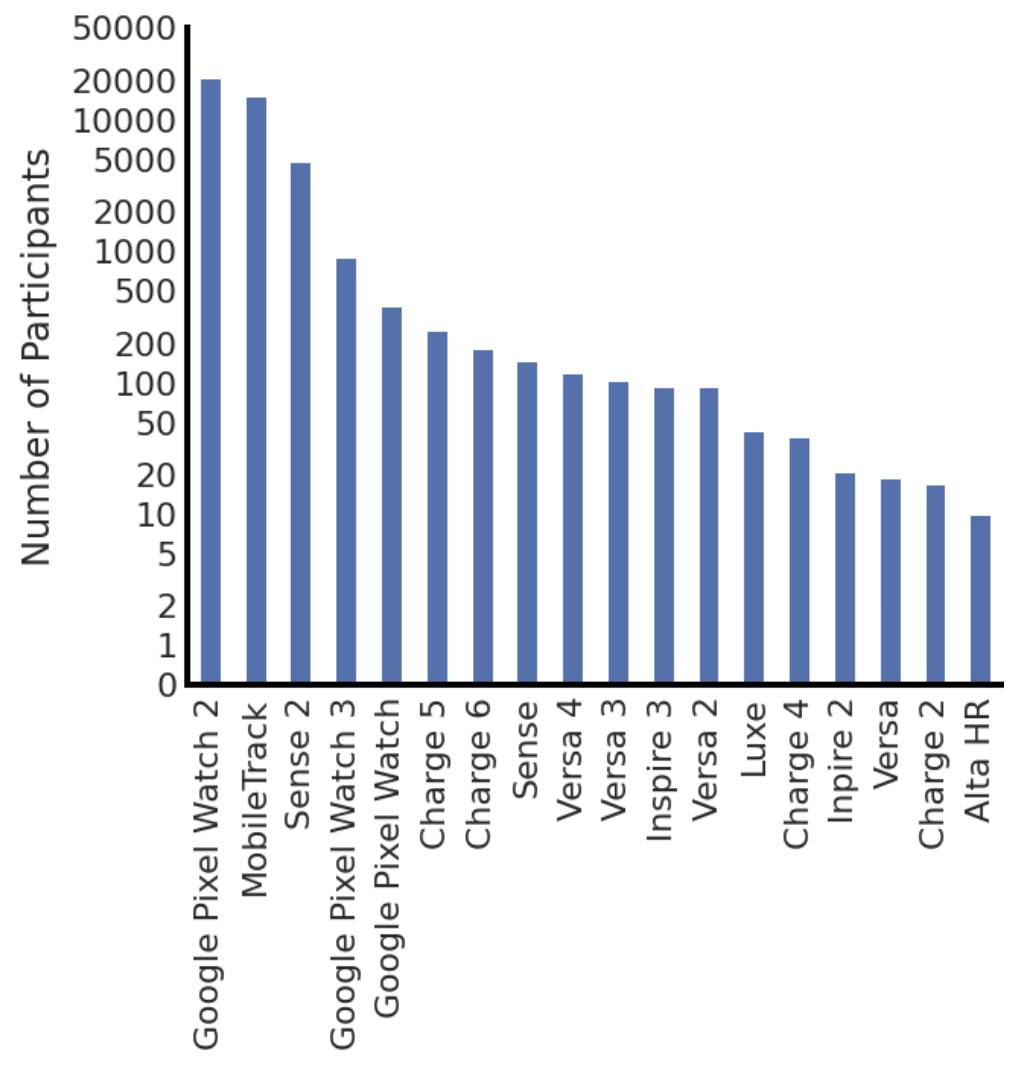}
    \caption{\textbf{Device Distribution.} The count of each fitness tracker present in our pre-training dataset.}
    \label{fig:device_numbers} 
\end{figure*}

There are many different types of smartwatches and fitness trackers.  Fig.~\ref{fig:device_numbers} shows the distribution of different trackers and smartwatches present in our pretraining dataset. Given the scale of our dataset we are able to train on examples of data from many different devices. Consequently, our model demonstrates robustness across diverse device types, handling their varying sensor technologies and differing inherent missingness patterns.

\subsection{Sensor Derived Minutely Features}
Our wearable devices utilize 5 different sensors: Photoplethysmography, Accelerometer, Skin Conductance (electrodermal activity or EDA), Temperature, and Altitude. Each of these sensors collects raw waveform signals at 100 Hz, 25 Hz, 200 Hz, 6 Hz, amd 10 Hz respectively, but we do not use the signals at this high resolution because (1) due to practical reasons (i.e. prohibitive storage costs and battery drain), data is not stored in this raw form at our scale, and (2) it is computationally impractical to learn models on raw waveforms across an entire day (i.e. 200 Hz for 1 day is $T=17$million time-points, per instance). As such, various features are curated from the raw waveforms as minutely aggregrated features and saved to be used as inputs into our model. Each of these features are grounded in the domain literature, based on prior work that has shown their clinical effectiveness. For example, heart rate variability metrics like RMSSD \citep{degiorgio2010rmssd} or Shannon Entropy of RR intervals \citep{afdala2017automatic} have well-established prognostic value for cardiovascular health, while accelerometry features like jerk ratio \citep{pan2020feasibility} effectively characterize movement quality.

Each of the derived features, as well as their base sensor origin, can be found in Table \ref{tab:features} below. For the targeted sensor removal experiments, as well as any other descriptions of the sensor as a whole, \emph{we refer to the sensor as all features derived from the sensor}. For example, when removing the PPG sensor in the targetted missingness experiment, we remove all PPG-derived features, from Heart Rate to Shannon Entropy RR Differences.

\begin{table}[!htbp]
\small
    \centering
    \caption{\textbf{Sensor Feature Definitions and the Sensor they are Derived From.}}
    \begin{tabular}{rlp{7cm}}
    \toprule
        \textbf{Feature}   & \textbf{Unit} & \textbf{Definition} \\
        \midrule
        \grayrow
        \multicolumn{1}{l}{\textbf{Photoplethysmography}} & & \\
        Heart Rate & Beats/Min & Mean of instantaneous heart rate. \\ 
        Heart Rate at Rest & Beats/Min & Mean of heart rate at rest.\\ 
        RR Percent Valid & \% & \% of 5-minute window with valid RR intervals.  \\ 
        RR 80$^{th}$ Percentile & Msec & 80$^{th}$ percentile of 5-minute window of RR ints. \\ 
        RR 20$^{th}$ Percentile & Msec & 20$^{th}$ percentile of RR ints. \\ 
        RR Median & Msec & Median RR interval. \\ 
        RMSSD & Msec & Root mean squared st. dev. of RR ints. \\ 
        SDNN & Msec & Standard deviation of RR intervals. \\ 
        Shannon Ent. RR & Nats & Shannon entropy of the RR intervals. \\ 
        Shannon Ent. RR Diffs  & Nats & Shannon entropy of the RR interval differences. \\ %

        \grayrow
        \multicolumn{1}{l}{\textbf{Accelerometer}} & & \\
        Step Count & Steps & Number of steps. \\ 
        Jerk Autocorrelation Ratio & a.u. & Ratio of lag=1 autocorrelation to energy in 1st 3-axis principal component.\\ 
        Log Energy & a.u. & Log of sum of 3-axis root mean squared magnitude. \\ 
        Covariance Condition & a.u. & Estimate of condition number for the 3-axis covariance. \\ 
        Log Energy Ratio & a.u. & Log of ratio of sum of energy in 1st 3-axis principal component over energy of 3-axis root mean squared magnitude.  \\ 
        Zero Crossing St.Dev. & Seconds & Standard deviation of time between zero crossing of 1st 3-axis principal component. \\ 
        Zero Crossing Average & Seconds & Mean of time between zero crossing of 1st 3-axis principal component.\\ 
        Axis Mean & a.u. & Mean of 3-axis\\ 
        Kurtosis & a.u. & Kurtosis of 3-axis root mean squared magnitude.\\
        Sleep Coefficient & a.u. & Sum of 3-axis max-min range with 16 log-scaled bins.\\
        \grayrow
        \multicolumn{1}{l}{\textbf{Skin Conductance}} & & \\
        Skin Conductance Value & $\mu$Siemens & Center of linear tonic SCL value fit. \\ 
        Skin Conductance Slope & $\mu$S/Min & Intraminute slope of SCL values. \\ 
        Lead Contact Counts & Counts & Number of times sensor leads contacted the wrist in a minute. \\ 
        \grayrow
        \multicolumn{1}{l}{\textbf{Skin Temperature}} & & \\
        Skin Temperature Value & $\degree$ C & Mean value of skin temperature. \\ 
        Skin Temperature Slope & $\degree$ C/Min & Slope of skin temperature. \\ 
        \grayrow
        \multicolumn{1}{l}{\textbf{Altimeter}} & & \\
        Altitude St.Dev. Norm & Hectopascals & Standard deviation of altimeter readings.\\ 
        \bottomrule
        \end{tabular}
    \label{tab:features}
\end{table}

\subsection{Demographic Breakdown}

A statistical breakdown of our datasets, by demographic features can be found in Table~\ref{tab:demographic_dataset_breakdown}. A subset of these, age and BMI, represent two of the regression tasks used to validate our method.

\begin{table}[!htbp]
\centering
\caption{\textbf{Demographics of our Various Datasets.}}
\label{tab:demographic_dataset_breakdown}
\resizebox{\textwidth}{!}{
\begin{tabular}{lrrrrrr}
\toprule
 & \multicolumn{2}{c}{\textbf{Pre-training}} & \multicolumn{2}{c}{\textbf{Downstream Activity}} & \multicolumn{2}{c}{\textbf{Downstream Metabolic}} \\
\textbf{Category} & \textbf{Train (\%)} & \textbf{Val (\%)} & \textbf{Train (\%)} & \textbf{Val (\%)} & \textbf{Train (\%)} & \textbf{Val (\%)} \\
\midrule
\textbf{Sex} \\
Male & 37,352 (68.1) & 3,657 (63.8) & 27,653 (73.1) & 6,092 (73.0) & 551 (44.1) & 258 (35.4) \\
Female & 23,041 (38.1) & 2,065 (36.0) & 10,145 (26.8) & 2,248 (26.9) & 670 (53.6) & 455 (62.4) \\
Not Specified & 48 (0.1) & 10 (0.2) & 24 (0.1) & 3 (0.1) & 0 (0) & 0 (0) \\
\midrule
\textbf{Age} \\
18--39 & 28,519 (47.2) & 2,583 (45.1) & 19,340 (51.1) & 4,492 (53.8) & 415 (33.2) & 223 (30.6) \\
40--59 & 24,888 (41.2) & 2,433 (42.4) & 15,309 (40.5) & 3,172 (38.0) & 637 (51.0) & 384 (52.7) \\
60--79 & 6,473 (10.7) & 664 (11.6) & 2,875 (7.6) & 618 (7.4) & 198 (15.8) & 121 (16.6) \\
$\geq$80 & 364 (0.6) & 39 (0.7) & 120 (0.3) & 31 (0.4) & 0 (0) & 1 (0.1) \\
Not Specified & 197 (0.3) & 178 (0.5) & 30 (0.4) & 0 (0) & 0 (0) & 0 (0) \\
\midrule
\textbf{BMI} \\
Healthy ($<$25) & 22,425 (37.1) & 2,173 (37.9) & 15,942 (42.2) & 3,685 (44.2) & 319 (25.5) & 188 (25.8) \\
Overweight (25--30) & 20,242 (33.5) & 1,952 (34.1) & 14,154 (37.4) & 3,017 (36.2) & 343 (27.4) & 206 (28.6) \\
Obese ($\geq$30) & 14,799 (24.5) & 1,330 (23.2) & 6,131 (16.2) & 1,316 (15.8) & 481 (38.5) & 274 (37.6) \\
Not Specified & 230 (0.4) & 14 (0.2) & 81 (0.2) & 18 (0.2) & 49 (3.9) & 28 (3.8) \\
\midrule
\textbf{Total} & 60,440 (100) & 5,732 (100) & 37,822 (100) & 8,343 (100) & 1,250 (100) & 729 (100) \\
\bottomrule
\end{tabular}
}
\end{table}

\subsection{Discriminative Task Label Breakdown}

Table~\ref{tab:discriminative_dataset_breakdown} shows label and data breakdown of the discriminative tasks used to validate our method. These tasks include 20-class activity recognition (Table~\ref{tab:discriminative_dataset_breakdown}(a)) from the activity dataset, and binary anxiety and hypertension classification (Table~\ref{tab:discriminative_dataset_breakdown}(b.i)) from the metabolic dataset.

\begin{table}[!htbp]
\setlength{\tabcolsep}{2.5pt}
\caption{\textbf{Discriminative Task Dataset Distribution}}
\label{tab:discriminative_dataset_breakdown}
\small
\begin{center}

\begin{minipage}[t]{0.49\textwidth}

\centering
\adjustbox{max width=\textwidth}{
\begin{tabular}{clrr}
\multicolumn{3}{l}{(a) \textbf{Activity Recognition Dataset}} \\
\toprule[1.5pt]
& \textbf{Task / Label} & \textbf{Train (\%)} & \textbf{Test (\%)}\\
\midrule\midrule
& \textbf{Activity} \\
\walkIcon & Walk & 4,434 (6.0) & 874 (5.8)\\
\bikeIcon & Bike & 4,363 (5.9) & 858 (5.6)\\
\sportsIcon & Sport & 4,433 (6.0) & 902 (5.9)\\
\runningIcon & Run & 4,023 (5.4) & 790 (5.2)\\
\aerobicsIcon & Aerobics & 4,417 (6.0) & 906 (6.0)\\
\ellipticalIcon & Elliptical & 4,402 (5.9) & 879 (5.8)\\
\spinningIcon & Spinning & 4,402 (5.9) & 858 (5.6)\\
\weightliftingIcon & Weightlifting & 4,335 (5.9) & 841 (5.5)\\
\swimIcon & Swim & 4,280 (5.7) & 867 (5.8)\\
\hikingIcon & Hike & 4,062 (5.5) & 841 (5.5)\\
\tennisIcon & Tennis & 4,138 (5.6) & 815 (5.4)\\
\crossfitIcon & CrossFit & 4,305 (5.8) & 887 (5.8)\\
\pilatesIcon & Pilates & 4,365 (5.9) & 846 (5.6)\\
\stairclimberIcon & Stairclimber & 4,272 (5.8) & 834 (5.5)\\
\dancingIcon & Dancing & 4,288 (5.8) & 826 (5.4)\\
\climbingIcon & Indoor climbing & 3,520 (4.8) & 853 (5.6)\\
\golfIcon & Golf & 3,003 (4.1) & 710 (4.7)\\
\skiingIcon & Skiing & 1,594 (2.1) & 420 (2.8)\\
\snowboardingIcon & Snowboarding & 662 (0.9) & 167 (1.1)\\
\kayakingIcon & Kayaking & 732 (1.0) & 212 (1.4)\\
\midrule
& \textbf{Total} & 74,030 (100) & 15,186 (100)\\
\bottomrule[1.5pt]
\end{tabular}}

\end{minipage}
\hfill
\begin{minipage}[t]{0.49\textwidth}
\vspace{-148pt}
\centering
\adjustbox{max width=\textwidth}{
\begin{tabular}{lcc}
\multicolumn{3}{l}{(b.i) \textbf{Metabolic Dataset} Classification Tasks} \\
\toprule[1.5pt]
\textbf{Task / Label} & \textbf{Train (\%)} & \textbf{Test (\%)}\\
\midrule\midrule
\textbf{Anxiety} \\
Positive & 55,030 (36.4) & 34,749 (38.5)  \\
Negative & 96,316 (63.6) & 55,437 (61.5) \\
\midrule
\textbf{Hypertension} \\
Positive & 36,349 (24.0) & 23,353 (25.9) \\
Negative & 114,997 (76.0) & 66,833 (74.1) \\
\midrule
\textbf{Total} & 151,346 (100) & 90,186 (100)\\
\bottomrule[1.5pt]
\\\\
\end{tabular}}

\end{minipage}
\end{center}
\end{table}

\subsection{Acquisition and Approval}

The data used for training in our analysis was curated from a large corpus of historical wearable data collected with consent from partcipants for these data to be used in research.  Specifically, the consent language described use of the data for developing new health features and algorithms and being included in publications:

\textit{REDACTED will collect and use your data to research and develop new health and wellness products and services for you and others. This data includes your: Health and wellness data, such as steps, heart rate, and sleep data.
Your data may also be used to generate findings that could be included in publications (such as scientific journals) to contribute to general knowledge about health and science. For example, activity, heart rate, and sleep data contributed to published findings that Fitbit devices could help detect flu outbreaks. None of the data used for these purposes will include your name, email, or other information that directly identifies you.}

The use of data for pretraining in this manner was approved as exempt under 45 CFR § 46.104(d)(4) \textit{"because the research involves the use of identifiable private information/biospecimens; and information, which may include information about biospecimens, is recorded by the investigator in such a manner that the identity of the human subjects cannot readily be ascertained directly or through identifiers linked to the subjects, the investigator does not contact the subjects, and the investigator will not re-identify subjects."}

The Metabolic downstream dataset for anxiety and hypertension prediction came from an IRB approved study (protocol number removed for anonymization). The core objective of this study as described in the IRB protocol was to: \textit{"Evaluate the feasibility of using the data provided by wrist-worn wearable devices to develop algorithms and scores to assess metabolic health."}

In the consent for the observational study, participants were informed that data on up to 7,500 participants in the United States would be collected. We used a mobile study platform that allows participants to enroll, check eligibility and provide full informed consent. The same mobile application enables the collection of Fitbit data using Fitbit devices or Pixel watches and allows participants to complete questionnaires. The participants reported their anxiety, depression and hypertension diagnoses through this app. Data was de-identified and stored in accordance with the approved IRB protocol. The participants were compensated with a free set of lab tests from Quest Diagnostics for participating in the study.

\section{Missingness Visualizations} \label{appendix:missdetail}

A core property of these data is that they are fragmented, and the missingness has several modal types. Three very common modes occur: 1) When the device is being charged or off all sensor stop recording data (device off), 2) when the device is in certain operation modes (e.g., when in sleep mode) certain signals stop being recorded (sensor off) and 3) when there is noise in the sensor data spurious values (e.g., values that are not physiologically possible - HR=0) are filtered out. The following sections demonstrate additional visualizations of the missingness patterns present from these mechanisms.

\subsection{Additional Examples of Data with Existing Missingness}

\begin{figure*}[!htbp]
    \centering
    \includegraphics[width=\textwidth]{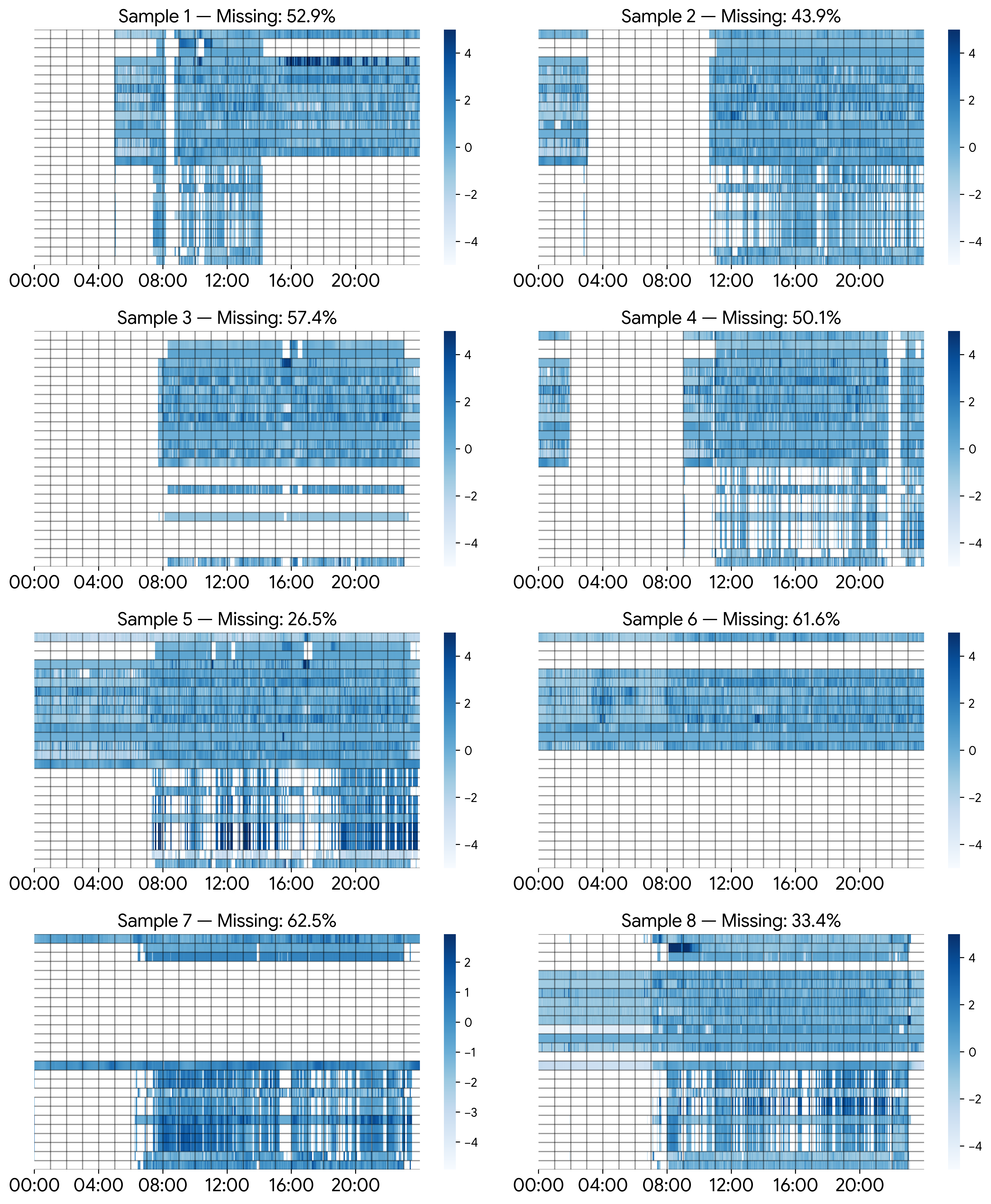}
    \caption{\textbf{Gallery of Data Examples with Real-world Missingness.} White designates missingness.}
    \label{fig:missexamples} 
\end{figure*}

In order to demonstrate the ubiquity and broad range of missingness patterns found within the data, we randomly sample an additional 8 data examples, shown in Figure \ref{fig:missexamples}. These examples further demonstrate how some patterns are consistent across users, such as increased missingness during early morning hours (12am-6am) (reflecting device removal during sleep) or correlated missingness dropout across various sensor channels. However, it should be noted that all samples exhibit unique missingness signatures with no two patterns being identical with vastly differing missingness percentages (27-63\%) and demonstrating the ubiquity of real-world missingness. These findings motivated our development of \projectname's flexible masking approach, which explicitly models such heterogeneous missingness patterns during pre-training.

\subsection{Prevalence and Length of Missingness}
\begin{figure*}[!htbp]
    \centering
    \includegraphics[width=\textwidth]{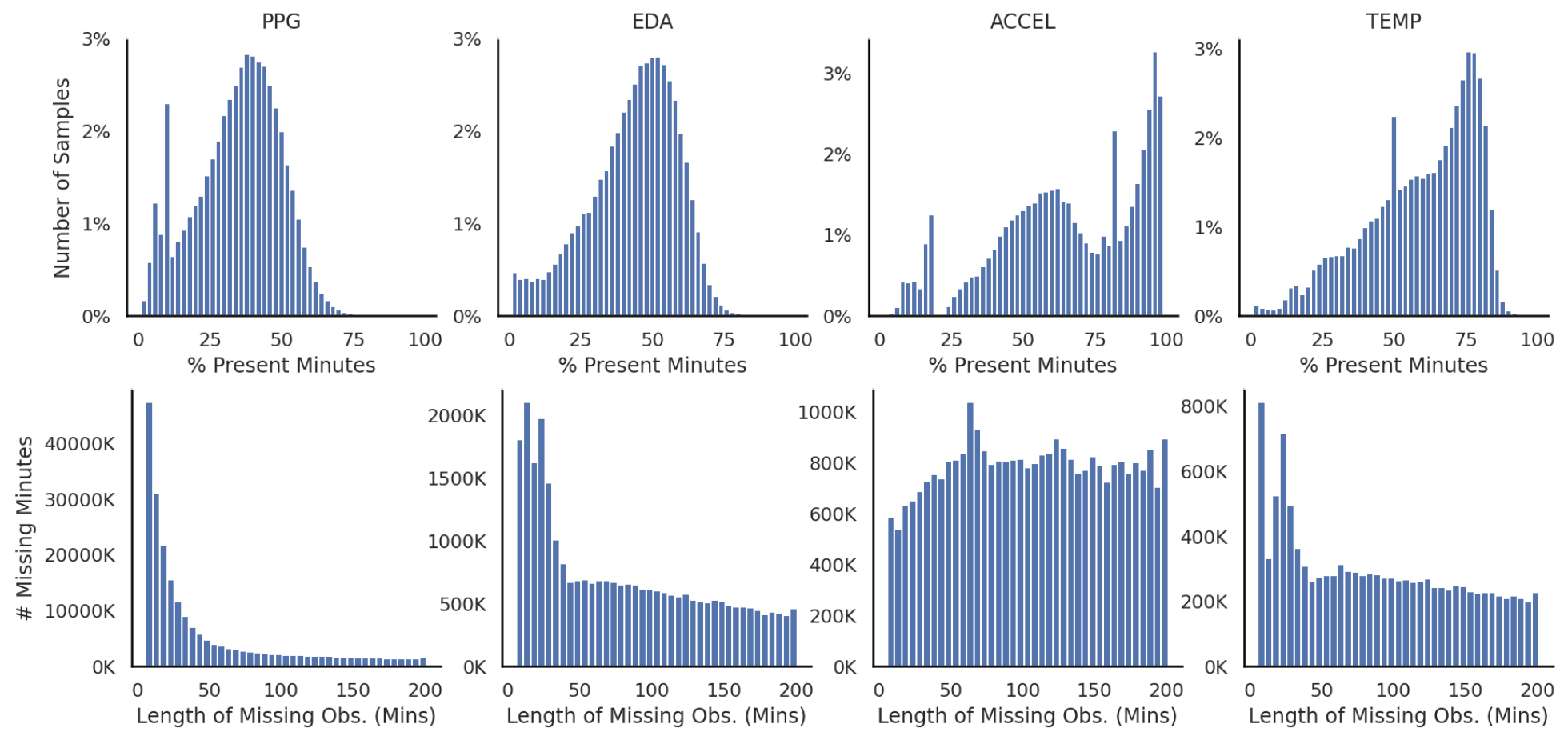}
    \caption{\textbf{Distribution of Prevalence and Length of Missingness.}}
    \label{fig:present_minutes} 
\end{figure*}

In Figure \ref{fig:present_minutes}, we demonstrate the prevalence of missingness as well as the length of the missingness, broken down across each sensor type across all 1.6 million instances of pre-training data. As we can see, each sensor has very different patterns of missingness, and across all sensors, their missingness presents as long extended gaps, making them non-trivial to reconstruct over. Notably, the accelerometry features in particular, have missingness in the form of these extended gaps, whereas most of the missingness for PPG sensors is of shorter length. 

\section{Pre-training Masking \% Ablation Experiment} \label{appendix:maskingablation}
The adaptive component of our \projectname methodologies allows for us to utilize a mix of artificial mask pre-training masking strategies. Each of these artificial masks are applied ontop of the existing, inherited mask.  In order to model both dimensionalities of our data, across time and sensors, and the real-world missingness paradigms, we have a mix of 3 different artificial mask pre-training strategies:
\begin{enumerate}
    \item Random Imputation Pre-training: Here we drop out a \% of total tokens. This is useful for modeling sensor noise, in which random channels at random times will be missing.
    \item Temporal Slice Pre-training: Here we drop out a \% of total temporal slices, across all sensor channels. This is useful for modeling device off, in which, for a given period of time, all sensors are off because the wearable device is off body. Here, we do not model it like temporal interpolation, in which the slices are necessarily contiguous. This is because, during pre-training, we would like to learn to reconstruction across a variable number of contiguous slices. 
    \item Sensor Slice Pre-training: Here we drop out a \% of total sensor slices, across all time points. This is useful for modeling sensor off, in which a given sensor channel is off because of a non-random missingness mechanism that tells the device to turn off the channel (i.e. to save battery life). 
\end{enumerate}

Below in Tables \ref{tab:ablaterimp}, \ref{tab:temporal-imputation}, \ref{tab:modality-imputation}, we see that an 80\% random imputation mask \%, 50\% temporal slice \%, and a 50\% sensor slice \% produce a good mix of reconstruction results across small and large amounts of evaluation masking, for each generative task. Note that when there is a tie, we would prefer higher masking \%, in order to allow for a higher dropout removal ratio, and to produce a harder task for our model to pre-train with.

\begin{table}[!htbp]
\centering
\caption{\textbf{Effect of Differing Pre-training Random Imputation Mask \% on Random Imputation.}}
\label{tab:ablaterimp}
\begin{tabular}{@{}lccc@{}}
\toprule
                     & \multicolumn{3}{c}{Random Imp. Eval Ratio} \\ \midrule
PT Random Imp. Mask \% & 30\%        & 50\%        & 80\%       \\ \midrule \midrule
90\%                & 0.13        & 0.14        & 0.20       \\
\grayrow
\textbf{80\%}                   & 0.10        & 0.12        & 0.19       \\
70\%                  & 0.10        & 0.12        & 0.19       \\
60\%                   & 0.10        & 0.12        & 0.19       \\
50\%                   & 0.09        & 0.12        & 0.20       \\ \bottomrule
\end{tabular}%
\end{table}

\begin{table}[!htbp]
\centering
\caption{\textbf{Effect of Differing Pre-training Temporal Slice Mask \% on Temporal Interpolation.}}
\label{tab:temporal-imputation}
\begin{tabular}{@{}lcccc@{}}
\toprule
                     & \multicolumn{4}{c}{Temporal Interp. Eval Amount} \\ 
\cmidrule(lr){2-5}
PT Temporal Slice \% & 10 min & 30 min & 60 min & 180 min \\ 
\midrule \midrule
70\% & 0.23 & 0.34 & 0.41 & 0.56 \\
60\% & 0.26 & 0.36 & 0.42 & 0.57 \\ 
\grayrow
\textbf{50\%} & 0.23 & 0.33 & 0.40 & 0.55 \\
40\% & 0.22 & 0.33 & 0.40 & 0.56 \\
30\% & 0.22 & 0.33 & 0.40 & 0.57 \\
\bottomrule
\end{tabular}
\end{table}

\begin{table}[!htbp]
\centering
\caption{\textbf{Effect of Differing Pre-training Sensor Slice Mask \% Ratios on Sensor Imputation.}}
\label{tab:modality-imputation}
\begin{tabular}{@{}lcccc@{}}
\toprule
                     & \multicolumn{4}{c}{Sensor Imp. Eval Amount} \\ 
\cmidrule(lr){2-5}
PT Sensor Slice \% & 2/26 & 6/26 & 12/26 & 24/26 \\ 
\midrule \midrule
70\% & 0.19 & 0.23 & 0.28 & 0.43 \\
60\% & 0.18 & 0.22 & 0.27 & 0.45 \\ 
\grayrow
\textbf{50\%} & 0.17 & 0.21 & 0.27 & 0.48 \\
40\% & 0.17 & 0.21 & 0.27 & 0.56 \\
30\% & 0.16 & 0.21 & 0.30 & 0.63 \\
\bottomrule
\end{tabular}
\end{table}

\clearpage
\section{Model Hyperparameter and Implementation Details} \label{appendix:hyperparams}
\subsection{Pre-training Set-up.} We pre-train our models on a large set of wearable minutely sensor data described. The raw multimodal sensor data input can be denoted by $\textbf{A} \in \mathbb{R}^{T \times S}$. $S=26$, which is the full number of signals in our multimodal data. These signals are derived from 4 different wearable sensors: Accelerometry, PPG, EDA, and Temperature. In our setting, we set $T=1440$, which is composed of all minutes from a full 24 hour day,  from midnight to midnight local time. We use this window size as days normally have a consistent structure, allowing for a more meaningful absolute positional embedding than if an arbitrary window size was set (e.g. 300 minutes  \citep{narayanswamy2024scaling}).

Our model was pre-trained with a ViT-1D \citep{dosovitskiy2020image, abbaspourazad2023large} encoder backbone by using a 1D patch size of 10 time-steps (i.e. 10 minutes). This results in a total of 3744 tokens (the 1440 minutes are reduced to 144 tokens per signal. With 26 signals, 26*144=3744 is the final number of tokens). Similar to prior work \citep{na2024guiding}, each signal channel is patched with a shared kernel, and we utilize a 2D positional embedding to encode information about the temporal position and signal channel. The ViT model had 25 million parameters with an encoding dimensionality of 384, 12 encoder layers, and 4 decoder layers. Our mask is a union of the inherited mask with an artificial masking mix of 80\% random imputation, 50\% temporal slices, and 50\% signal slices. Our primary pre-training objective is to optimize the signal reconstruction loss (i.e. mean squared error), averaged over the artificially masked patches. The model was pre-trained on 8x16 Google v5e TPUs with a total batch size of 512 across 100,000 training steps. The training process uses the AdamW optimizer with a base learning rate of $5e-3$, weight decay set to $1e-4$, and betas set to $0.9$ and $0.95$. Gradients were clipped at $1.0$. A linear warm-up schedule is applied for the first 5\% of total steps, followed by a cosine learning rate decay to zero.

Our SSL baselines include LSM \citep{narayanswamy2024scaling}, SimCLR \citep{chen2020simple}, DINO \citep{caron2021emerging}, and a Masked Siamese Network (MSN) \citep{assran2022masked}. LSM is an MAE \citep{he2022maskedmae} approach with 0.8 random masking ratio with no inherited masking. SimCLR, DINO, and MSN are augmentation-based contrastive approaches, and we utilize a set of common time-series augmentations \citep{tang2020exploring, liu2024guidelines, zhang2022self, rommel2022data}:  jittering, scaling, and time flipping. Each augmentation has a 0.5 probability of being applied. Jittering was implemented as a random sample from a gaussian distribution with zero-mean and a uniformly randomly sampled standard deviation frp, 0 to 0.5, per value in the time-series. Scaling was implemented by multiplying all of the data input with a scale, uniformly sampled from 1.1 to 1.5. For DINO, we omit scaling as the model was unable to converge.

Each of these baselines were all pre-trained from scratch, following the same previously stated training conditions, unless stated otherwise. All baselines expect full, complete data as input, and as such, they utilize the imputed version of our sensor dataset. LSM was trained with a ViT-2D with a 2D patch size of (10,2), in order to match their image-based encoding approach, and all other ViT parameters remain constant.

\subsection{Downstream Evaluation}
We group our downstream evaluation into three sections based on the target: generative, classification, and regression.

In our \textbf{Generative Evaluation}, we evaluate how well our model is able to reconstruct different types of structured missingness patterns that mimic real-world missingness patterns: (1) Random Imputation, where a [30\%, 50\%, 80\%] of tokens is masked out, (2) Temporal Interpolation, where all signals in a contiguous temporal window of length [10, 30, 60 minutes] is completely masked out, (3) Temporal Extrapolation, which is similar to interpolation, but the window is necessary at the end of the time-series, and (4) Signal Imputation, where all time points for a random set of [2/26, 6/26, 12/26] signal channels is masked. Reconstruction performance was calculated with mean squared error (MSE) on the artificially masked tokens, averaging only over the data points that have a ground truth.

Our deep learning baselines include the LSM model \citep{narayanswamy2024scaling}, another MAE-based model, which can be used to evaluate these generative tasks out-of-box by setting the artificial masking procedure to match the proposed tasks. Our \projectname model is done in the same way, but the full encoder mask includes the inherited mask as well. Unfortunately, the contrastive SSL baselines are unable to provide generative performance metrics because they do not utilize a reconstruction objective. Instead, we use alternative simple generative baselines, which match practical applications. Many application-focused biosensor algorithms will employ simple imputation methods \citep{pires2020improving, xu2022pulseimpute, srimedha2022comprehensive, wu2020deep, amiri2016missing} as quick data preprocessing methods. Thus, we choose to include these additional methods as baselines: Linear Interpolation, K-Nearest Neigbhors, and Mean Filling. Similar to our method, we run these baselines with a union mask of the mask inherited from existing missingness and the artificial mask. MICE \citep{van2011mice} is another popular, simple baseline designed for multivariate data, but we opted to not include it due to our existing missingness patterns violating the Missingness At Random assumption, and prior work demonstrate a relative poorer performance compared to nearest neighbor and linear interpolation \citep{narayanswamy2024scaling}.

In our \textbf{Classification Evaluation}, we evaluate how well our model's embedding representation is able to capture discriminative features. During evaluation, our model calculates the embedding on all non-inherited-masked tokens and uses an average pooling followed by a trainable linear probe to classify each of the prediction targets. For the LSM model, because it is unable to represent the inherited mask, the embedding for all tokens is pooled, such that tokens that were part of the existing missingness but have been filled with imputation will be included. For the contrastive methods, the learned CLS token is used as the pooled representation. We report performance with F1 score as it balances precision and recall for class-imbalanced targets, Accuracy as a straightforward measure of overall correctness, Balanced Accuracy to account for potential class imbalance, and AUROC to evaluate the model's ranking capability across all classification thresholds. The prediction targets are hypertension, anxiety, which originate from the Metabolics dataset and 20-class activity recognition, which originates from the Activity dataset. 

The linear probe was trained by freezing the learned ViT backbone, averaging over the entire embedding and training a logistic regression head ontop of it. For our \projectname model specifically, with the inherited mask, the average was only done over the non-masked tokens. Training was done with a batch size of 512, across 500 training steps with an AdamW optimizer with a base learning rate of $5e-3$, weight decay set to $1e-4$, and betas set to $0.9$ and $0.95$. Gradients were clipped at $1.0$. For activity specifically, training steps and learning rate were increased to 1000 and $1e-1$ to achieve better convergence. 

Additionally, we include two extra supervised baselines, ViT-1D \citep{dosovitskiy2020image} and a ResNet \citep{he2016deep}, that are trained end-to-end for each of our tasks. ViT-1D is a transformer-based architecture that follows the same architecture as our \projectname with 25 million parameters, but with randomly initialized weights, trained end-to-end. ResNet is a strong CNN-based architecture that has seen broad success throughout the health biosignal time-series domain \citep{xu2024relcon, pillai2024papagei, abbaspourazad2023large, mekruksavanich2022resnet}. This model was a ResNet-50 \citep{he2016deep} with 25 million parameters, in order to match the ViT model. Specifically, it contains 50 layers, with 64 filters that double after each residual block, with a final average pooling and logistic regression head. Both models are trained with a batch size of 512, across 500 training steps with an AdamW optimizer with a base learning rate of $5e-3$, weight decay set to $1e-4$, and betas set to $0.9$ and $0.95$. Gradients were clipped at $1.0$. A linear warm-up schedule is applied for the first 5\% of total steps, followed by a cosine learning rate decay to zero. Because these models do not handle missingness, they were trained directly on the imputed data.

In our \textbf{Regression Evaluation}, we utilize the same evaluation procedure described in classification, only instead the linear probe is specifically a linear regression. We report performance with MAE as it provides an interpretable deviation from the correct value, as well as Pearson Correlation Coeffecient, as it is a common metric for evaluating how well a regressor is able to capture the trend of the target \citep{xu2024relcon, yuan2024self}. The prediction targets are BMI and Age. 

The linear probe was trained by freezing the learned ViT backbone, averaging over the entire embedding and fit a linear regression head ontop of it using Scikit-Learn's LinearRegression implementation out-of-box. The supervised baselines were trained in an identical way as done in the classification evaluation, but using a linear regression head instead of logistic regression.

\clearpage
\section{Additional Results}
\subsection{Confusion Matrices}
\begin{figure*}[!htbp]
    \centering
    \includegraphics[width=\textwidth]{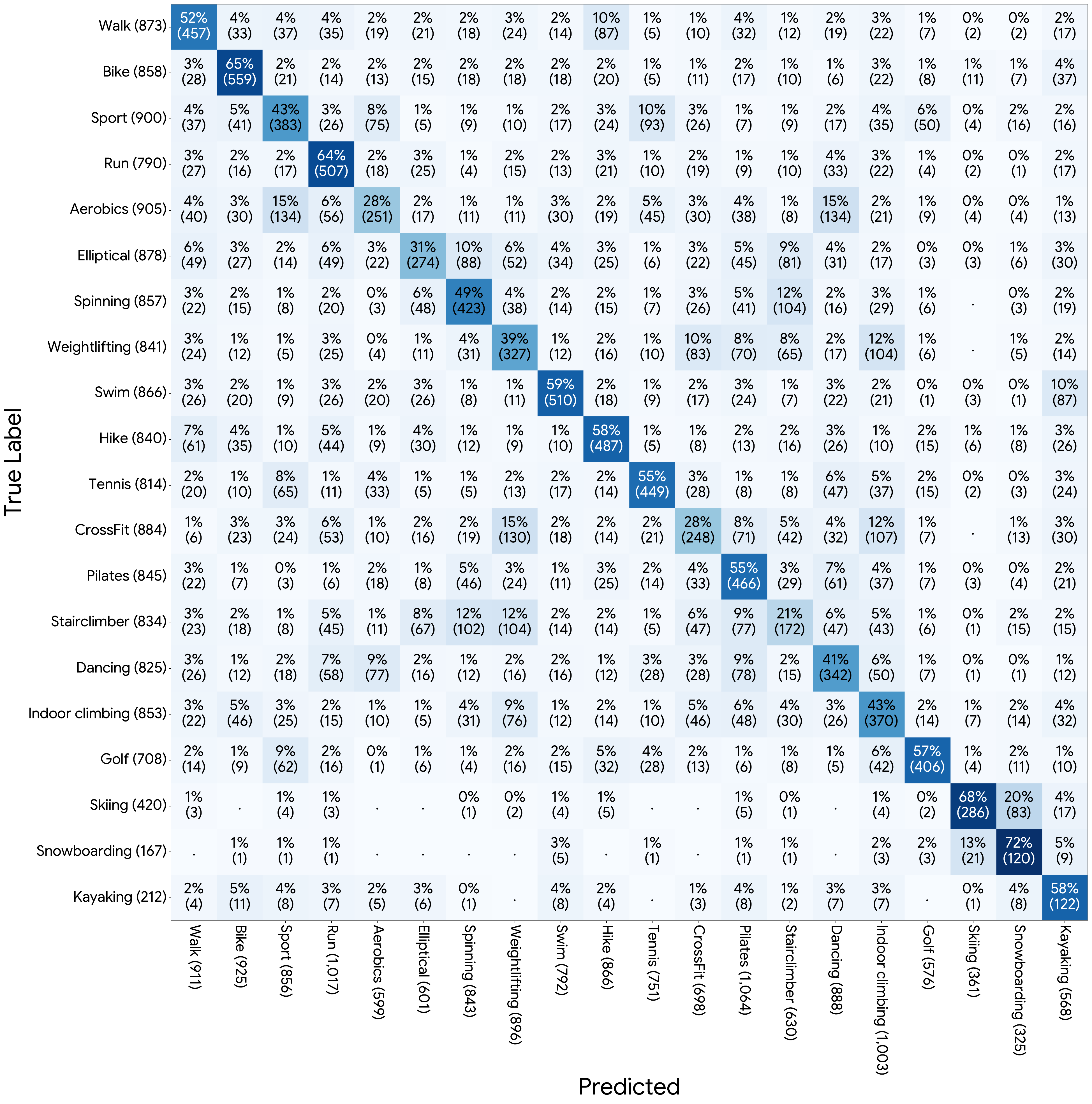}
    \caption{\textbf{Activity Recognition Confusion Matrix.} The results of a linear probe applied to \projectname for the 20-class activity recognition task. Rows add up to 100\%.}
    \label{fig:cf_mtx} 
\end{figure*}

Figure~\ref{fig:cf_mtx} illustrates the utility of \projectname learned embeddings for downstream applications. Specifically, this confusion matrix shows the performance of \projectname, post-trained on the 20-class activity recognition task using a linear probe. It is clear that the embedding are useful in discriminating between a large number of activities, even those which may be semantically clustered, such as skiing and snowboarding. Future work may explore how to expand to even more activities and behavioral events, and investigate the utility of large-scale pre-training in address long-tail task labels.

\subsection{Reconstruction Examples}
\begin{figure*}[!htbp]
    \centering
    \includegraphics[width=\textwidth]{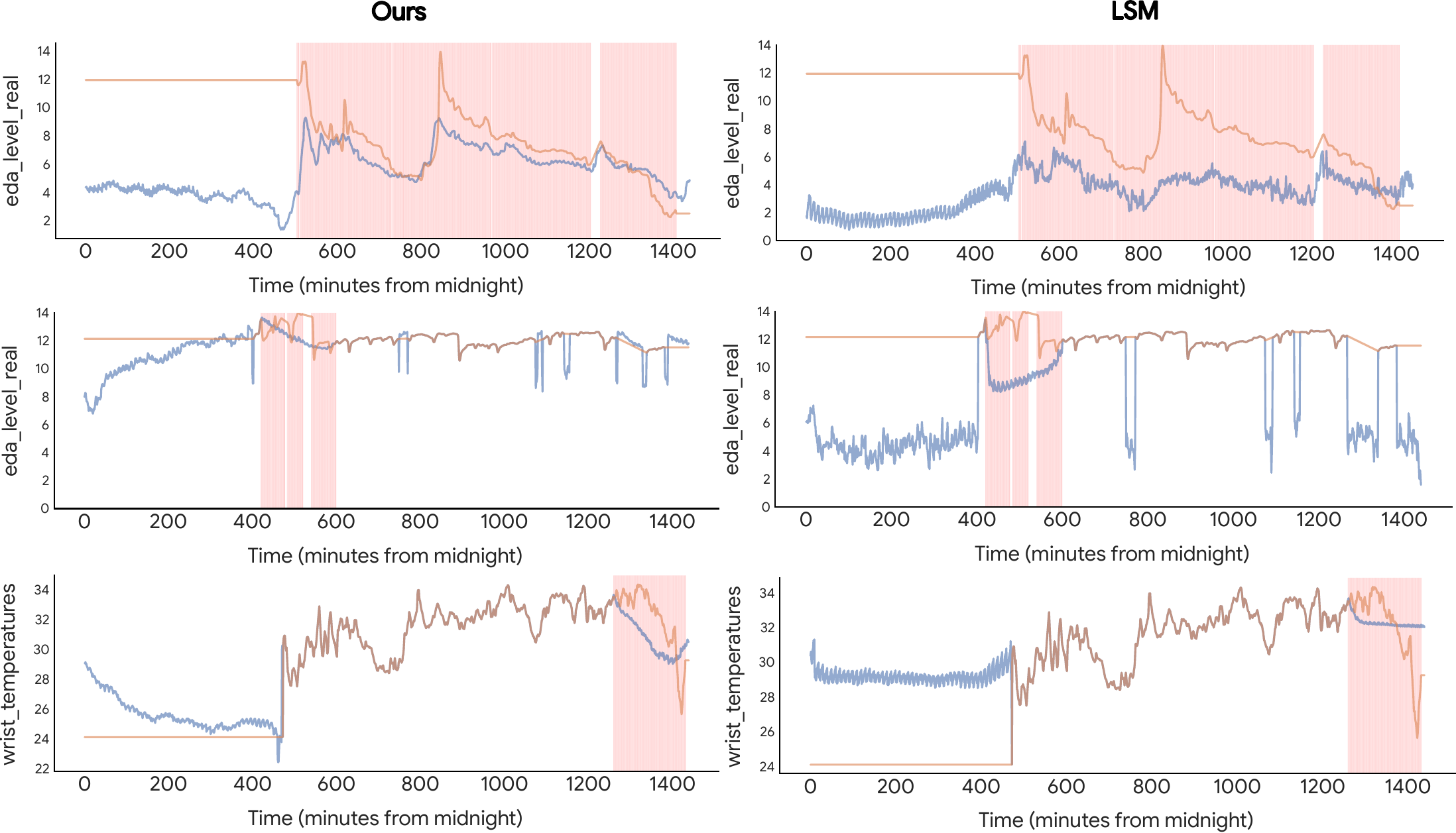}
    \caption{\textbf{Reconstruction Examples for 2/26 Sensor Signal Imputation (Row 1), 3 Hour Temporal Interpolation (Row 2), 3 Hour Temporal Extrapolation (Row 3).} Red highlighted regions demonstrate regions of artificial masking. Orange shows original data with imputation (i.e. the first 400-500 steps of the each row were originally missing, then imputed, as demonstrated by the straight line) and blue shows the reconstructed data.}
    \label{fig:reconex} 
\end{figure*}

Figure \ref{fig:reconex} shows various reconstruction examples for a specific sensor signal. Here we can clearly see Our \projectname approach leads to much stronger performance, across different generative tasks.

\section{Additional Discussions}

\subsection{The Utility of Day-Level Features} Traditionally, generalist methods for time-series health signals have focused on small windowed segments of data on the order of seconds or sub-seconds~\citep{abbaspourazad2023large, xu2024relcon, narayanswamy2024bigsmall, yuan2024self}. Such methods allow for fine-grain activity and physiological tracking. An adjacent body of work has explored the utility of longer observations, on the order of hours~\citep{spathis2021self, narayanswamy2024scaling}, enabling more complex person-level insights. In this work seek to expand the observation window to encode a high-level of context. Day level features allow models to learn relationships not possible from shorter spans, for example, how a person's activity during the day may affect their night-time resting heart rate. Looking forward, we intend to continue exploring how best to encode large context windows to include known week, seasonal, and year level periodicities.

\subsection{Person-Level versus Event-Level Performance} Analysis of the discriminative results (classification and regression) presented in the main body of the paper, raise an interesting question: how do generative pre-training affect performance on person-level and event-level tasks. For person-level tasks (hypertension, anxiety, age, BMI) we find that \projectname consistently outperforms supervised baselines while only using a simple linear probe. In contrast, we find for the event-level task (20-class activity recognition), ResNet50, a supervised baseline performs extremely well, and likely a fully-finetuned \projectname model is needed to surpass it. This suggest that while supervised methods easily capture event-level features (e.g., sudden heart rate changes due to activity), they struggle to learn slow-changing, near-constant day-level features more-relevant to person-level tasks. This highlights how method, like are own, learn a more complex representation of the data via generative pre-training. We further concede that our contrastive SSL baselines fail to fully realize the gains of pre-training. We hypothesizes that more complex time-series augmentations are needed to leverage their effect.

\subsection{Limitations and Future Work}

Here we expand upon the limitations and future work introduced in the main body of the paper.

\textbf{Generalizing to New Devices.} Though many commodity wearables host a similar suite of sensors there are inevitable differences between these software-hardware systems. We acknowledge that our methods focuses on a small subset of such devices. Future work will explore the generalizability of our methods to additional devices and datasets, and investigate the extent to which device specific missingness patterns result in a distribution shift.

\textbf{Generalizing to Open Data.} Most publicly available wearable datasets (e.g. WESAD \citep{schmidt2018introducing}, PAMAP2 \citep{bleser2015personalized}) are composed of high-frequency raw signals that are very limited in their temporal context with only a subset of the sensors we have available. Thus, they are unable to shown to be used in our setting of day-level context. All of Us \citep{Jeong2025allofus} demonstrates an interesting avenue to apply our work. Although limited to only the Heart Rate and Step Count channels (compared to our 26 channels), the dataset contains with long context windows and minutely data, and presents an interesting direction in future work to apply our \projectname method. 

\textbf{Data and Feature Scales.} Time-series analysis often requires explicit assumptions regarding data scale. As such, our method focuses on day-long samples. We acknowledge that such data disregards known periodicities (e.g., weekly, seasonal, etc.). Future work will explore combining our fine-grained behavioral and physiological modeling with insights from longer windows. Furthermore, our method utilizes minutely aggregated features as opposed to the raw sensor feeds common in sensing research. This is a practical limitation, as data is not stored in its raw form at this scale.

\textbf{Handling Sensor Feature.} Our method utilizes 26 features derived from a set of 5 sensors, and regards each feature as independent in the modeling. In reality there are significant correlations between features from the same sensor (e.g., heart rate and heart rate variability). More work can be done to explore how best to combine these multimodal features -- potentially sensor-specific encoders, cross-attention, or special class tokens per-sensor feed.

\subsection{Broader Impact}
Personal and ubiquitous health technologies, including smart phones and wearables, have the potential to scale to billions of individuals. Such devices allow for significant self- and longitudinal tracking, and in so doing may augment the current paradigm of clinical healthcare. To-date, consumer health technologies focus on low-level insights, such as steps, resting heart-rate, and sleep staging, which allow users to reason on personal higher-level insights (e.g., "my resting heart-rate has been elevated ever since I fell sick").

In contrast, our method, trained on day-level samples, learns behavioral and physiological patterns useful in deriving more complex insights. For example, our method shows the potential to predict anxiety and hypertension, insights that humans and commercial algorithms would struggle to derive given only sensor data. We believe this line of work will one day enable people to make the most of their tracked wearable data, better understand their behavior and physiology, and in so doing receive more proactive and better informed care.

\subsection{Ethics Statement}
While consumer health research holds potential for significant positive impact, with so many possible stake holders, such research must be performed intentionally to ensure that it is safe and fair. Additionally, there exists the unfortunate possibility that bad-actor may attempt to leverage methods, such as our own, in negligent ways. As researchers in the field, the burden falls to us to consider the implications of this research, and act to fulfill the positive impacts and mitigate the associated risks.

Building upon this, we concede that training our methods on closed (non-public) data, prevents the scientific community from fully replicating our work. We acknowledge this as a limitation and attest our support for open science and open data. However, due to the sensitive nature of health data, these considerations must be balanced by with the privacy and protection of our participants.


\clearpage

\end{document}